\documentclass{article}

\usepackage[preprint, nonatbib]{neurips_data_2024}
\usepackage[utf8]{inputenc} % allow utf-8 input
\usepackage[T1]{fontenc}    % use 8-bit T1 fonts
\usepackage{hyperref}       % hyperlinks
\usepackage{url}            % simple URL typesetting
\usepackage{booktabs}       % professional-quality tables
\usepackage{amsfonts}       % blackboard math symbols
\usepackage{nicefrac}       % compact symbols for 1/2, etc.
\usepackage{microtype}      % microtypography
\usepackage{xcolor}         % colors
\usepackage{graphicx}
\usepackage{wrapfig}
\usepackage{subcaption}
\usepackage{svg}
\usepackage{amsmath} 
\usepackage[capitalize]{cleveref}
\usepackage{verbatim}
\usepackage{enumitem}   
\usepackage{marvosym}
\usepackage{tcolorbox}
\usepackage{listings}
\newcommand{\eg}{\emph{e.g.}}

\title{MMtrail: A Multimodal Trailer Video Dataset with Language and Music Descriptions}

\author{%
Xiaowei Chi$^{1}$\footnotemark[1] , Yatian Wang$^{1}$\footnotemark[1],  Aosong Cheng$^{2}$\footnotemark[1] , Pengjun Fang$^{1}$\footnotemark[1] ,Zeyue Tian$^{1}$\footnotemark[1] ,\\
\textbf{Yingqing He$^{1}$, Zhaoyang Liu$^{1}$, Xingqun Qi$^{1}$, Jiahao Pan$^{1}$, Rongyu Zhang$^{2}$,}\\
\textbf{Mengfei Li$^{1}$, Ruibin Yuan$^{1}$, Yanbing Jiang$^{1}$, Wei Xue$^{1}$, Wenhan Luo$^{1}$, Qifeng Chen$^{1}$,}\\
\textbf{Shanghang Zhang$^{2 \textrm{\Letter}}$, Qifeng Liu$^{1 \textrm{\Letter}}$, Yike Guo$^{1}$}\\
$^{1}$ The Hong Kong University of Science and Technology \\
$^{2}$ Peking University \\
}

\begin{document}{
\maketitle

\footnotetext[1]{These authors contributed equally to this work.}
% \footnotetext[2]{Corresponding authors.}
\footnotetext[2]{$\textrm{\Letter}$ Corresponding authors.}
\footnotetext[3]{Github repository:\hyperlink{https://github.com/litwellchi/MMTrail}{https://github.com/litwellchi/MMTrail}}
\footnotetext[4]{Project Page:\hyperlink{https://mattie-e.github.io/MMTrail/}{https://mattie-e.github.io/MMTrail/}}
% \vspace{-0.5em}

% \vspace{-0.5em}
\begin{abstract}
\vspace{-0.5em}
Massive multi-modality datasets play a significant role in facilitating the success of large video-language models. 
However, current video-language datasets primarily provide text descriptions for visual frames, considering audio to be weakly related information. They usually overlook exploring the potential of inherent audio-visual correlation, leading to monotonous annotation within each modality instead of comprehensive and precise descriptions. Such ignorance results in the difficulty of multiple cross-modality studies.
To fulfill this gap, we present MMTrail, a large-scale multi-modality video-language dataset incorporating more than 20M trailer clips with visual captions, and 2M high-quality clips with multimodal captions. 
Trailers preview full-length video works and integrate context, visual frames, and background music. In particular, the trailer has two main advantages: 
(1) the topics are diverse, and the content characters are of various types, \eg, film,  news, and gaming.
(2) the corresponding background music is custom-designed, making it more coherent with the visual context. 
Upon these insights, we propose a systemic captioning framework, achieving various modality annotations with more than 27.1k hours of trailer videos. Here, to ensure the caption retains music perspective while preserving the authority of visual context, we leverage the advanced LLM to merge all annotations adaptively.
In this fashion, our MMtrail dataset potentially paves the path for fine-grained large multimodal-language model training. In experiments, we provide evaluation metrics and benchmark results on our dataset, demonstrating the high quality of our annotation and its effectiveness for model training. 
%We hope this dataset will contribute to the development of visual-audio understanding and its downstream tasks.

\end{abstract}   

    \begin{figure}[ht]
    \centering
    \includegraphics[width=1.0\linewidth ]{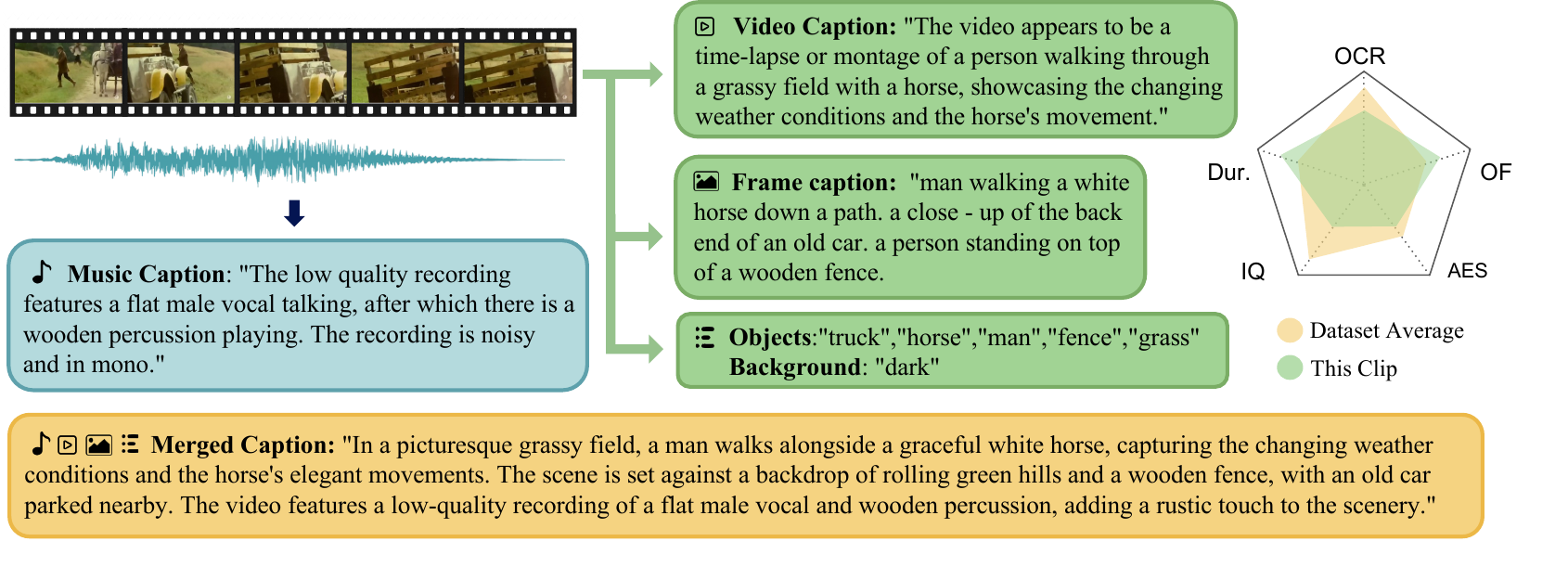}
    \caption{We present a video-language dataset with music captions, \textbf{MMTrail}.}
    % from an in-the-wild trailer video that contains multi-modality annotation and evaluation.}
    \label{fig:teaser}
\end{figure}}

\section{Introduction}
\label{sec:intro}
AI-driven movies and shot video production have a wide range of applications in people's daily lives.
Clearly, creating vivid videos requires more than just visual frame generation or individual modality-based ones. 
Thanks to various large-scale video-language datasets, numerous generative multimodal large language models have been developed to achieve this goal~\cite{svd,long2024videodrafter,chen2023videocrafter1,animate-a-story,lin2023videodirectorgpt,kondratyuk2023videopoet,wang2023gen,henschel2024streamingt2v}.
However, the existing video-language datasets~\cite{panda70m, HowTo100M, InternVid} typically focus on visual-based text descriptions, and they overlook the significance of the inherent visual-audio dependencies.
It presents a complex challenge that demands cohesive integration of multiple modalities, yet remains largely unexplored.

% TODO 怎么合理 highlight music
% 实际上，制造这样的多模态数据集依旧有许多困难
% Collecting high-quality hybrid multimodal datasets poses significant challenges, making it difficult to ensure consistency between different modalities. 
%Unlike previous video datasets~\cite{Webvid}, multimodal datasets with captions involve more complex data formats and substantial costs, further complicating the process.
Collecting high-quality multi-modality source data that preserves consistency between different modalities is challenging. 
Unlike previous datasets that only provide visual frame-based caption~\cite{Webvid}, multimodal datasets contain complex data formats (\eg, music), resulting in more labor-intensive and time-consuming costs in data processing and annotation.
Moreover, achieving a high correlation between audio and visual content presents challenges.

% general background 
% 预告片是优质的视频+多模态生成或者理解的数据源头
\begin{wrapfigure}{r}{0.5\textwidth}
    \centering
    \vspace{-0.6cm}
\includegraphics[width=0.5\textwidth]{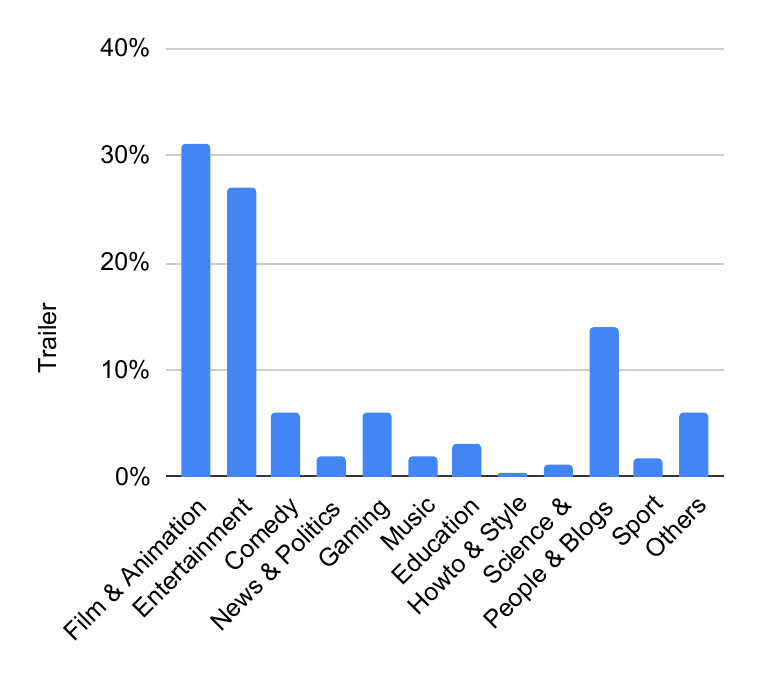}
\caption{Distribution of video categories of MMTrail dataset.}
\vspace{-0.5cm}
    \label{fig:distribute}
\end{wrapfigure}
Targeting to fill the dataset gap by creating a comprehensive and accurate multi-modality visual-audio dataset, we first notice trailers. As a precursor to a full-length work, the video trailer has emerged as a vital tool for artists to showcase and disseminate their creations. These short videos typically combine the most compelling visual shots with carefully selected music, have high cross-modality consistency, and hold significant potential in broader multimodal research. The topics are diverse, and the content characters are of various types, e.g., film, comedy, and gaming, as shown in ~\cref{fig:distribute}. Significantly, the trailer format represents a unique, high-quality, video-centric multimodal data source that benefits further multi-modality research exploration and analysis.

% 所以咱们的工作解决了这些困难
% 1. 弄了个大数据集
In this work, we propose \textbf{MMTrail}, which aims to unlock the potential of multimodal content understanding and generation for innovative applications in video content generation. We first recognize the immense value of trailers as a video-centric dataset, especially considering the music alongside the videos. \textbf{MMTrail} contains 20M+ video clips from 290k trailer videos encompassing various source categories as shown in \cref{fig:distribute}. To ensure the quality of our dataset, we have carefully designed a robust data filtering and cleaning methodology. We also provide extensive statistics works to demonstrate the diversity and complexity of our dataset.

% 2.我们提出了多模态fine-grande caption
To address the multimodal to language annotation challenge, we have designed a multimodal captioning pipeline incorporating diverse state-of-the-art (SOTA) captioning models~\cite{musiccap, Yu2022CoCa, liu2024llavanext}. Furthermore, we propose a language model fusion strategy to generate fine-grained multimodal captions. We have performed small-scale annotations on the entire dataset, created a multimodal annotation subset of 3 million samples \textbf{MMTrail-2M}, and provided a testing set \textbf{MMTrail-Test} with manually-adjusted multimodal caption.

% 3.我们在生成和理解任务上验证了这个数据集的优渥性能
We present evaluation metrics and benchmark results on our dataset, demonstrating the high quality of our annotations and their effectiveness for model training. Through extensive experiments and benchmarking, we showcase the difficulty and diversity of our dataset using various evaluation metrics. We also conduct human evaluations to validate the quality of our multimodal captioning pipeline. Furthermore, we fine-tune understanding models~\cite{videollama} and generative models~\cite{chen2023videocrafter1} on a subset of our dataset, providing evidence of its high quality and efficacy. Additionally, we evaluate video understanding models on the MMTrail-Test, highlighting the challenges posed by our dataset, and evaluate video-music-based models to demonstrate the effectiveness of cross-modality tasks.

% We hope this dataset will contribute to developing visual-audio understanding and its downstream tasks.
\begin{table}[t]
  \caption{Comparison of MMTrail-X and other Video to language datasets. MMTrail-X contains three sets(20M,2M, test) with 720p resolution. }
  \label{tab:videodataset}
  \footnotesize
% \scriptsize
\resizebox{\textwidth}{!}{%
  \centering
  \begin{tabular}{lcccccccc}
  \toprule
    Dataset     &Year  & Size & Caption & Modality &Clips & E(V) & E(T)  & Resolution  \\
    \midrule \midrule
    WebVid~\cite{Webvid} &  2021  & 52khr & Alt-text & Video & 10M & 10s & - & 360p    \\
    Panda~\cite{panda70m} &  2024  & 167khr & Auto & Video & 70M & 8.5s & 13.2  & 720p    \\
    HD-VILA~\cite{HD-VILA} &  2022  & 371.5khr & ASR & Video & 100M & 3.6s & 32.5 & 720p   \\
    MSR-VTT~\cite{MSR-VTT} &  2016  & 40hr & Manual & Video & 10K & 15s & 9.3 & 240p    \\ 
    InternVid~\cite{InternVid} &  2023  & 760.3khr & Auto & MM & 100M & 11.7s & 11.6 & 720p     \\
    HowTo100M~\cite{HowTo100M} &  2023  & 134.5khr & ASR & MM & 136M & 3.6s & 4 & 720p  \\
    \midrule
    \textbf{MMTrail-20M}     &  2024  &  27.1khr & Auto & Video & 20M & 4.6s & 10.7 & 720p  \\
    \textbf{MMTrail-2M}     &  2024  &  8.2khr & Auto & MM & 2M & 13.8s & 39.4 & 720p  \\
    \textbf{MMTrail-Test}     &  2024  &  3.2hr & Manual & MM & 1k & 11.6s & 98.2 & 720p  \\
    \bottomrule
  \end{tabular}
  }
\end{table}

\section{Related Work}
\label{sec:rw}

\vspace{-0.5em}
\subsection{Video generation and understanding}
Video understanding and text-to-video generation are inherently connected tasks. In recent years, there has been remarkable progress in understanding models~\cite{wu2021star,liu2021tam,zhao2022towards,bain2021frozen,yang2022zero,yang2022learning,lin2022swinbert,lin2019tsm,wu2019long,videollama,chen2023video}, which have greatly contributed to the advancement of text-based video generation techniques. The availability of large-scale datasets and diffusion models has revolutionized video generation, moving from pixel-level approaches like~\cite{vdm, singer2022make-a-video, ho2022imagen-video} to latent-level video diffusion models~\cite{he2022lvdm, zhou2022magicvideo, blattmann2023align, animate-a-story}. Concurrently, understanding models have also witnessed significant improvements. A series of MLLM-based understanding models~\cite{liu2024llavanext,maaz2023videochatgpt,song2023moviechat,jin2023chatunivi,jin2023chatunivi} has reach a satisfied understanding abilities. Besides, some surveys give detailed summaries of this area from different application areas or aspects~\cite{he2024llmsmeetmultimodalgeneration}. The iterative interaction between video generation and understanding has led to the development of excellent large-scale datasets and models encompassing diverse approaches. Panda~\cite{panda70m} introduced an auto-caption model distilled from video understanding models like VideoLlaMA~\cite{videollama}, MiniGPT4~\cite{zhu2023minigpt4}. Furthermore, leading to a large amount of multimodal generation models~\cite{lin2023videodirectorgpt,xiang2024pandora, chi2024m2chat, li2024mgm}.

\begin{figure}[t]
    \centering
    \includegraphics[width=1\linewidth ]{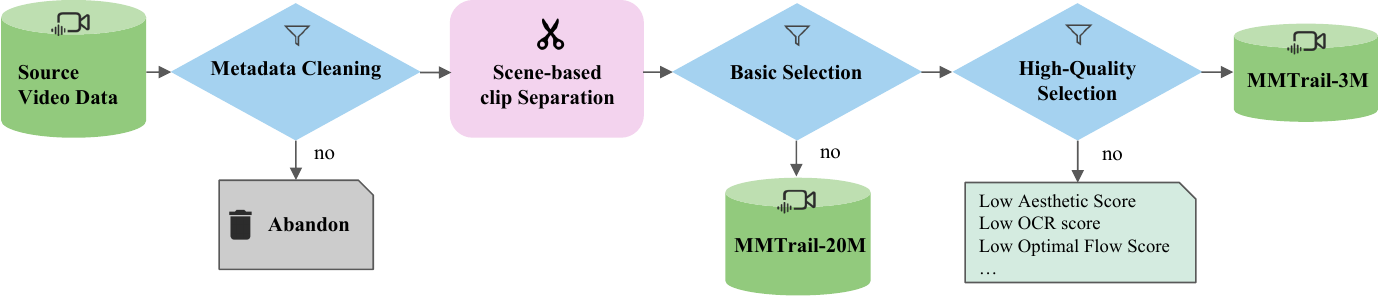}
    \caption{Data collection and cleaning pipeline of the MMTrail. Starting from the source video data, we follow the metadata cleaning, scene-cut, and basic filtering to obtain the full list of MMTrail-20M and High-Quality Selection to filter the MMTrail-2M.}
    \label{fig:clean_pipeline}
    % \vspace{-0.5em}
\end{figure}

\vspace{-0.5em}
\subsection{Video-Language datasets}

Captioned video datasets are essential for text-to-video generation and understanding tasks. MSR-VTT~\cite{MSR-VTT}, UCF-101~\cite{soomro2012ucf101} are commonly used as evaluation sets. Anna et al.~\cite{LSMDC} presented 118,081 movie clips with descriptions. ActivityNet Caption~\cite{ActivityNetCaptions} by Ranjay et al. is a benchmark involving event detection, natural language description, and event localization. WebVid~\cite{Webvid}, VideoFactory~\cite{HD-VG-130M}, and other works~\cite{How2, VaTeX, WTS70M, VideoCC3M}, contain multilingual video descriptions, video clips, metadata such as titles, descriptions, tags, and channel names, and are used for tasks like video understanding, text-to-video retrieval, and audio-video captioning with weak annotations. Several datasets and approaches have utilized audio to enhance video captioning~\cite{HowTo100M, LSMDC, YT-Temporal-180M, InternVid,panda70m, HD-VILA}. These datasets consist of movies, web videos, YouTube videos, and high-resolution videos from popular YouTube categories, providing transcribed audio descriptions, narrations~\cite{HowTo100M}, ASR transcriptions~\cite{YT-Temporal-180M}, and multiple captions generated through a auto caption model~\cite{HowTo100M, panda70m, InternVid}.

However, existing works only use audio to enhance the video caption, using metadata or ASR to provide extra information for the video caption. A large-scale video-centric dataset that cooperates with high-quality music and multimodal caption is still lacking.

\section{MMTrail-20M Dataset}
\label{sec:method}
To boost the performance of multimodal generation tasks, we construct a high-quality multimodal dataset based on trailers, which offers a wealth of multimodal information and diverse categories that distinguish them from existing large-scale video-language datasets.

% We devised an automated data annotation and cleaning pipeline to construct the source data for trailers in \cref{sec:method-data}. We performed extensive rating and statistical analysis on the video materials. Additionally, we developed a multimodal captioning pipeline that generated rich multimodal captions for video segments in \cref{sec:method-ca}. Subsequently, leveraging the rich annotations obtained through the processes above, we carefully selected a series of high-quality subsets suitable for different multimodal tasks in \cref{sec:method-subset}. These subsets were curated based on specific criteria, ensuring their applicability to diverse multimodal tasks within the dataset.

\begin{figure}[t]
  \centering
  % \vspace{-6mm}
  \begin{subfigure}{0.48\linewidth}
      \includegraphics[width=\linewidth]{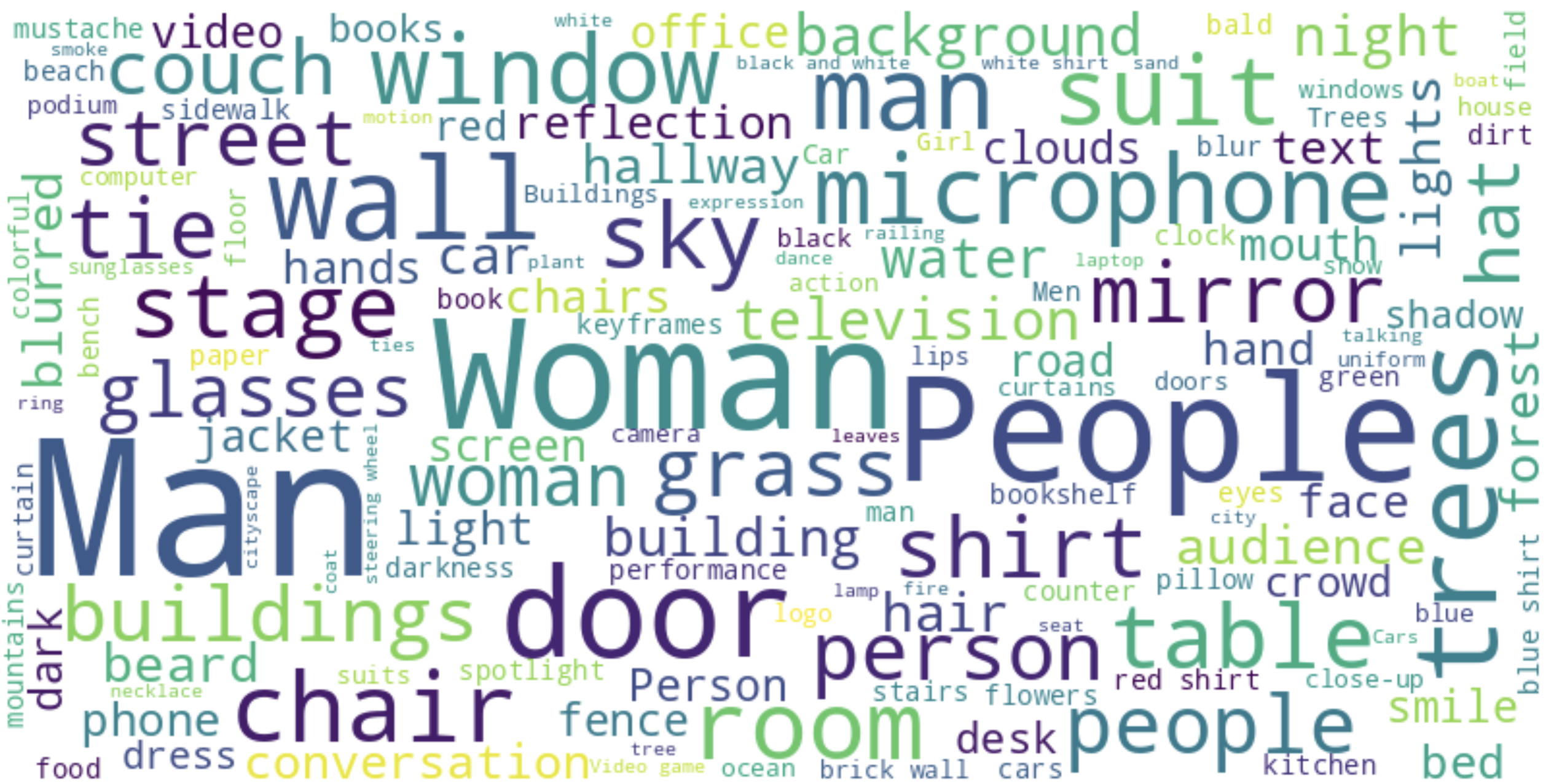}          
      % \caption{Word cloud of the objects.}
  \end{subfigure}
  % \hfill
          \hspace{0.02\textwidth}
  \begin{subfigure}{0.48\linewidth}
      \includegraphics[width=1.\linewidth]{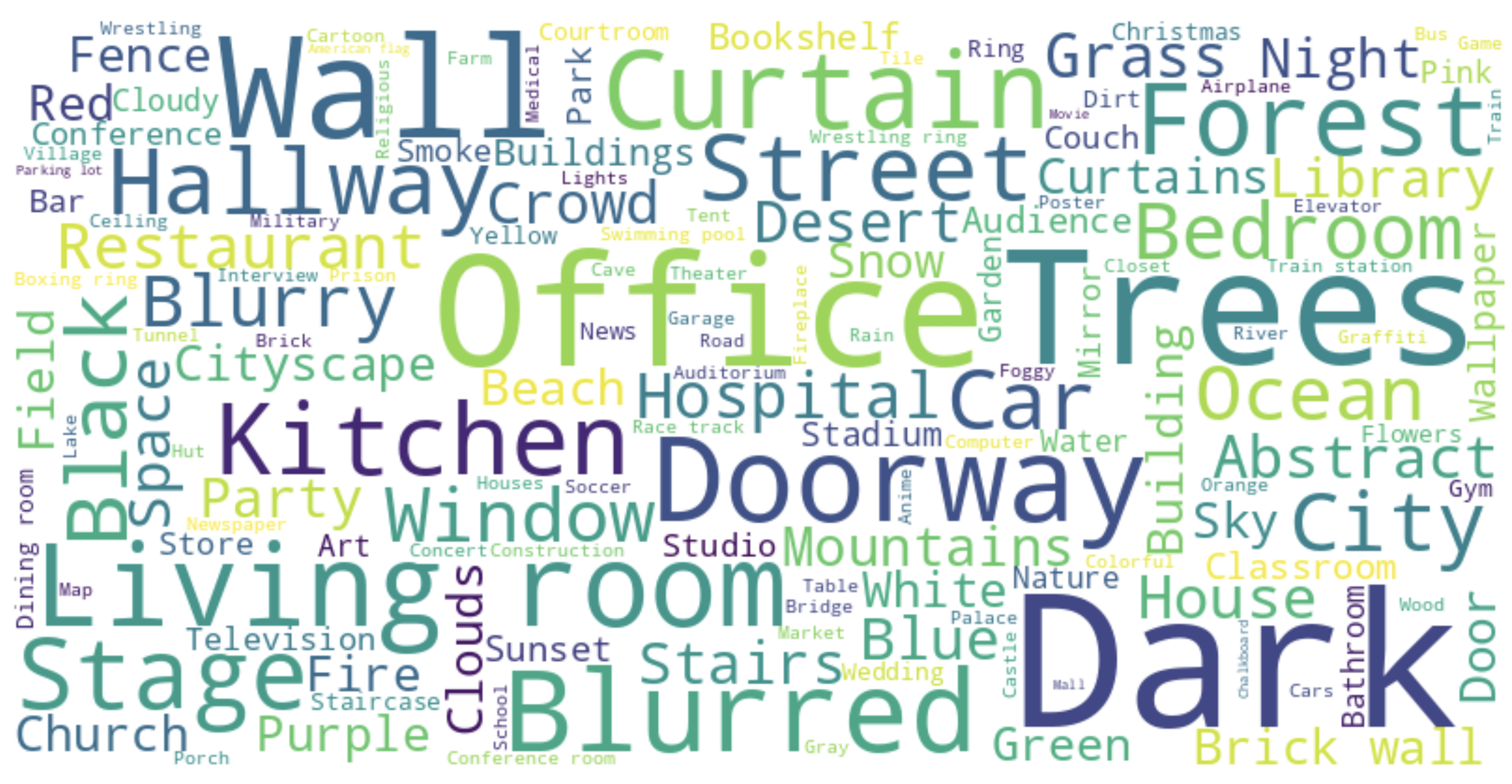}
  \end{subfigure}
  % \hfill
    \caption{Word cloud of the (left) objects and (right) background in MMTrail. Most of the objects are human, and most of the backgrounds are indoor scenes like office, kitchen, etc.}
  \label{fig:word_cloud}
  % \vspace{-0.5cm}
\end{figure}

\subsection{Data Collection Pipeline}
\label{sec:method-data}
Trailers, as unique, well-made video types, have particular data distribution and exceptional quality. The trailer videos usually contain multiple themes, including movies, TV shows, games, etc. They are well organized with the most attractive clips of the corresponding videos. At the same time, most of them are accompanied by corresponding background voices and music. Such rich and high-density information is a complete challenge for the current Artificial Intelligence Generation Content(AIGC) task and brings considerable difficulties to data cleaning. To solve this challenging problem, we designed a comprehensive data collection and cleaning process as shown in Fig.~\ref{fig:clean_pipeline} to deal with such complex videos, as described in this section.

\begin{figure}[t]
  \centering
 \includegraphics[width=\linewidth]{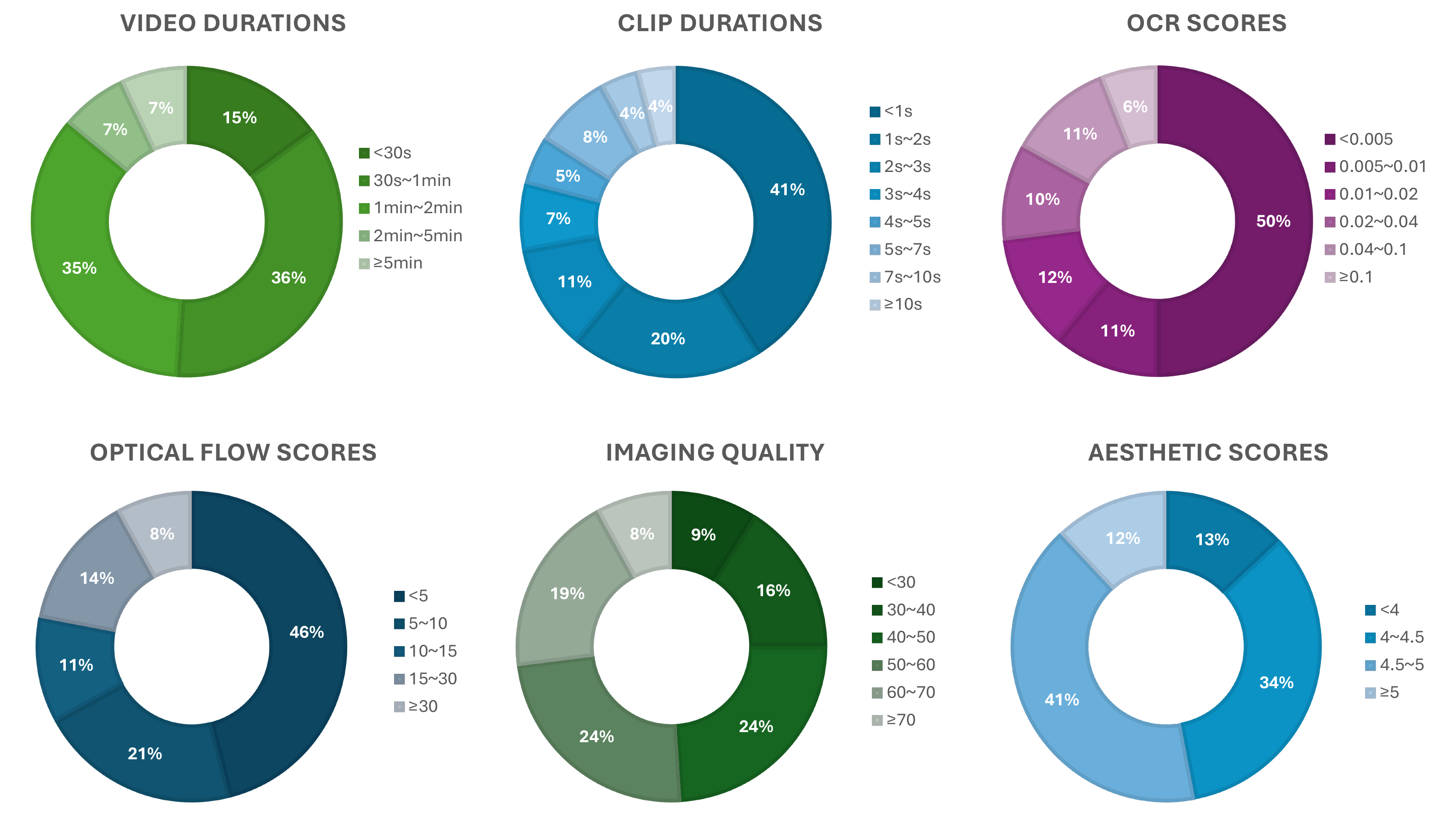}
  \caption{Statistic of the MMTrail clips. These evaluation scores collectively include OCR score, Video duration, optical flow score, clip duration, image quality, and aesthetic score, demonstrating the richness and diversity of MMTrail, making it a valuable resource for multimedia research.}
  \label{fig:statistic_all}
\end{figure}

\textbf{Collection Strategies}
To enrich the diversity of our MMTrail, we first employ the keyword ``trailer'' to retrieve various in-the-wild trailer videos. Such general-purpose trailer videos encapsulate a wide range of artistic works and genres. Then, we tailor the keywords more specifically to explicitly collect trailer videos with divergent sources. Those keywords include ``Movie Trailers'', ``Video Game Trailers'', ``TV Show Trailers'', ``Documentary Trailers'', etc. Through these two methods, we collect 285,518 comprehensive trailer videos with a total duration of 94,911,802.8 seconds.

\textbf{Trimming}
To facilitate the extraction of various video information in subsequent analyses, we cut the original videos into clips based on the scenes. As the most mature and practical tool currently available, PySceneDetect\footnote{https://github.com/Breakthrough/PySceneDetect} offers robust functionalities for this purpose. Thus, we use ContentDetector of PySceneDetect to compare the difference in content between adjacent frames and then cut the videos according to the predetermined threshold of 30. We finally generated 21,588,792 clips with an average duration of 4.6 seconds. 

\textbf{Motion Filtering} 
Evaluating motion quality is a crucial step in selecting high-quality video content. Motion vectors and optical flow are both mainstream motion quality evaluation methods. In our case, trailer videos often have rapid cuts and transitions between scenes, posing challenges for optical flow-based analysis. Other than optical flow, motion vectors are more robust to these quick changes, as they rely on larger block units' displacement rather than individual pixels' continuous flow. On the other hand, motion vectors are more lightweight and can be calculated more efficiently. Thus, we leverage motion vectors to filter out clips with problems like static frames, title sequences, and slideshow-like playback.
\begin{figure}[htb]
    \centering
    \begin{tabular}{cc|cc}
        \includegraphics[width=0.23\textwidth]{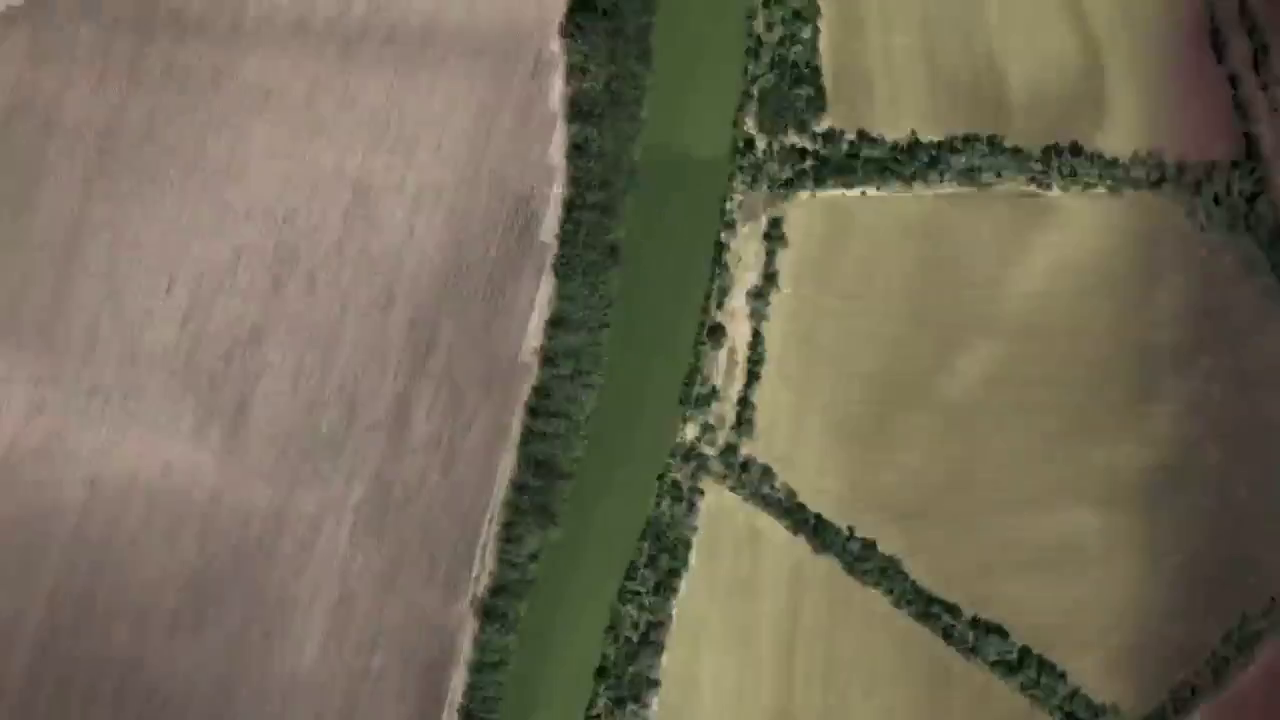} &
        \includegraphics[width=0.23\textwidth]{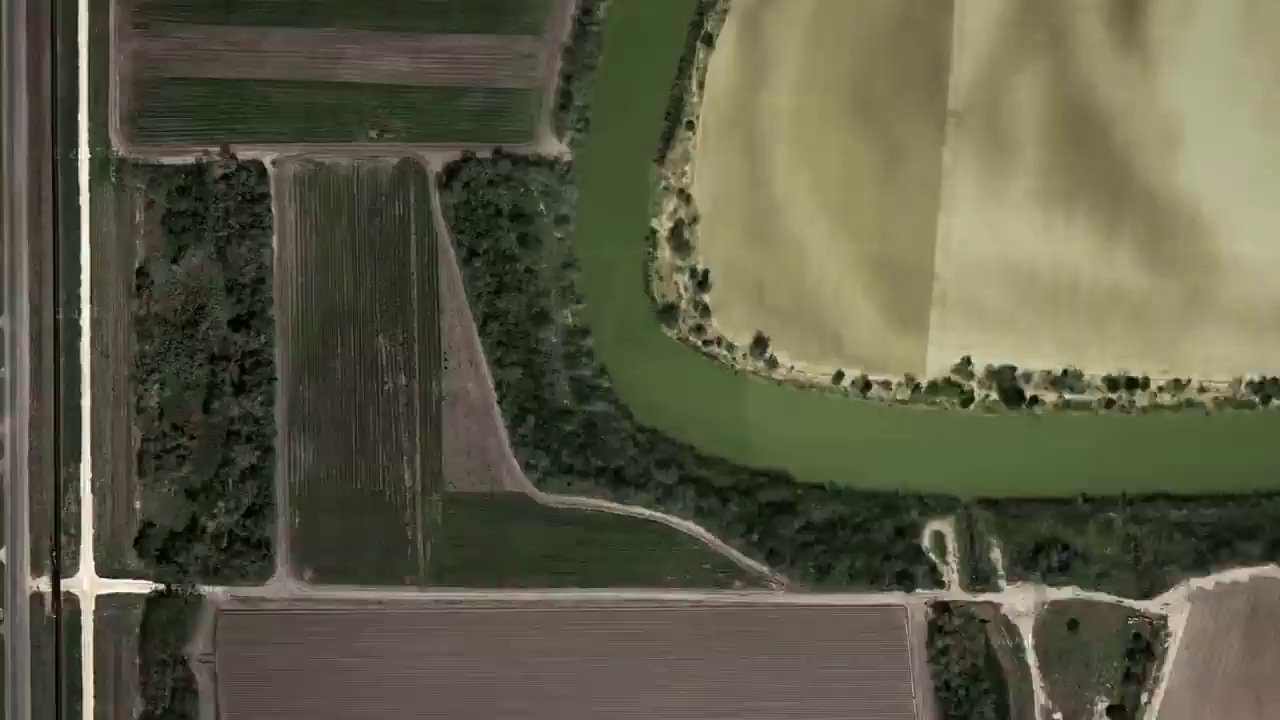} &
        \includegraphics[width=0.23\textwidth]{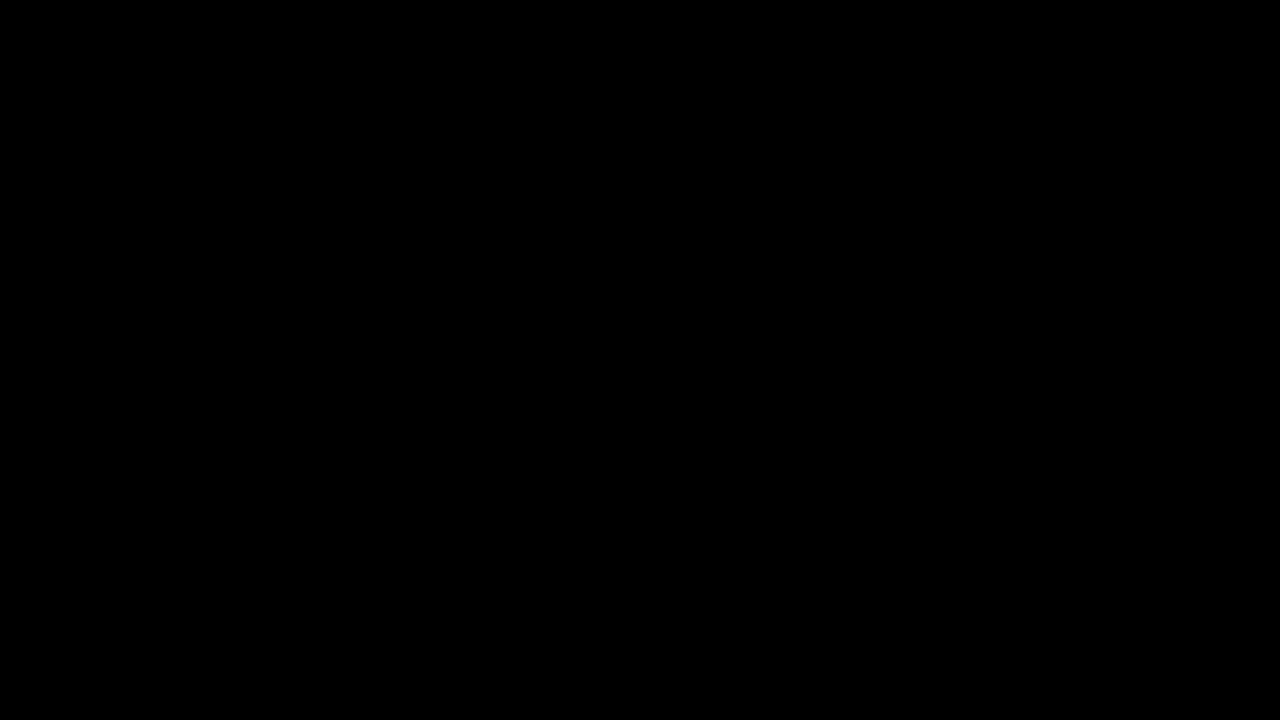} &
        \includegraphics[width=0.23\textwidth]{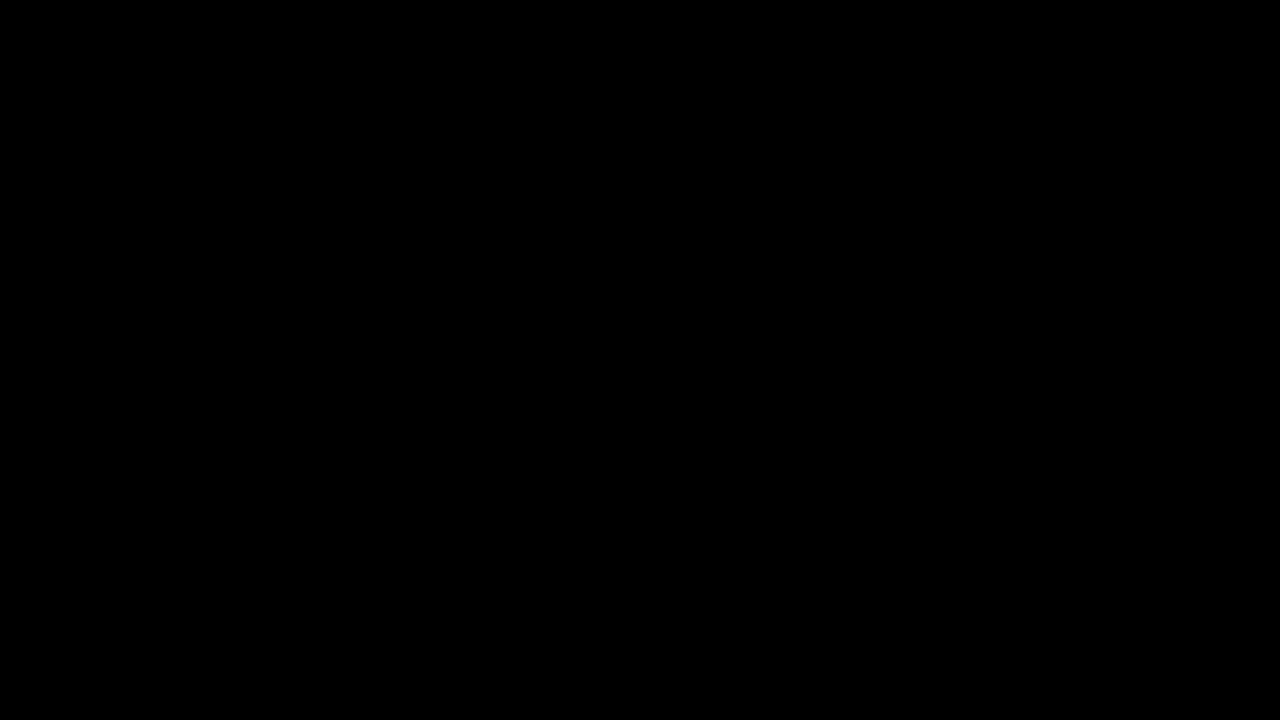} 
    \end{tabular}
    \caption{Clips of high (left) and low (right) motion quality scores. }
  \label{fig:motion_scores}
  % \vspace{-0.5cm}
\end{figure}

\textbf{Diversity}
We evaluate the diversity and richness of our dataset from three aspects: theme, objects, and backgrounds. While collecting, we first assess the categories from the Yotoube metadata provided by the video provider, as shown in ~\cref{fig:distribute}. Furthermore, we generate an object-level caption list and background by Llava-NexT~\cite{liu2024llavanext} for a more accurate category-based generation. The word cloud of objects and backgrounds is shown in ~\cref{fig:word_cloud}.

% \cxw{
% \textbf{Audio Video Alignment}
% We use ImageBind to assess the semantic alignment between the vision and audio modalities. The model is pre-trained in a CLIP fashion to align six different modalities. ImageBind-AV~\cite{girdhar2023imagebind} scores typically indicate a stronger semantic correlation between the vision and audio modalities. We compute the ImageBind-AV scores for all the data to evaluate this alignment. 
% }
% \cxw{ImageBind distribution Map}

\textbf{OCR}
% 这里可以用公式示例
Trailer videos often have text-heavy sections with high-quality text animations, like opening and ending credits. To identify these text-rich segments, we utilize OCR to detect the text content in the video frames and calculate the bounding box area of the text. This measurement reflects the amount of text in the clips. 
\begin{figure}[htb]
    \centering
    \begin{tabular}{cc|cc}
        \includegraphics[width=0.23\textwidth]{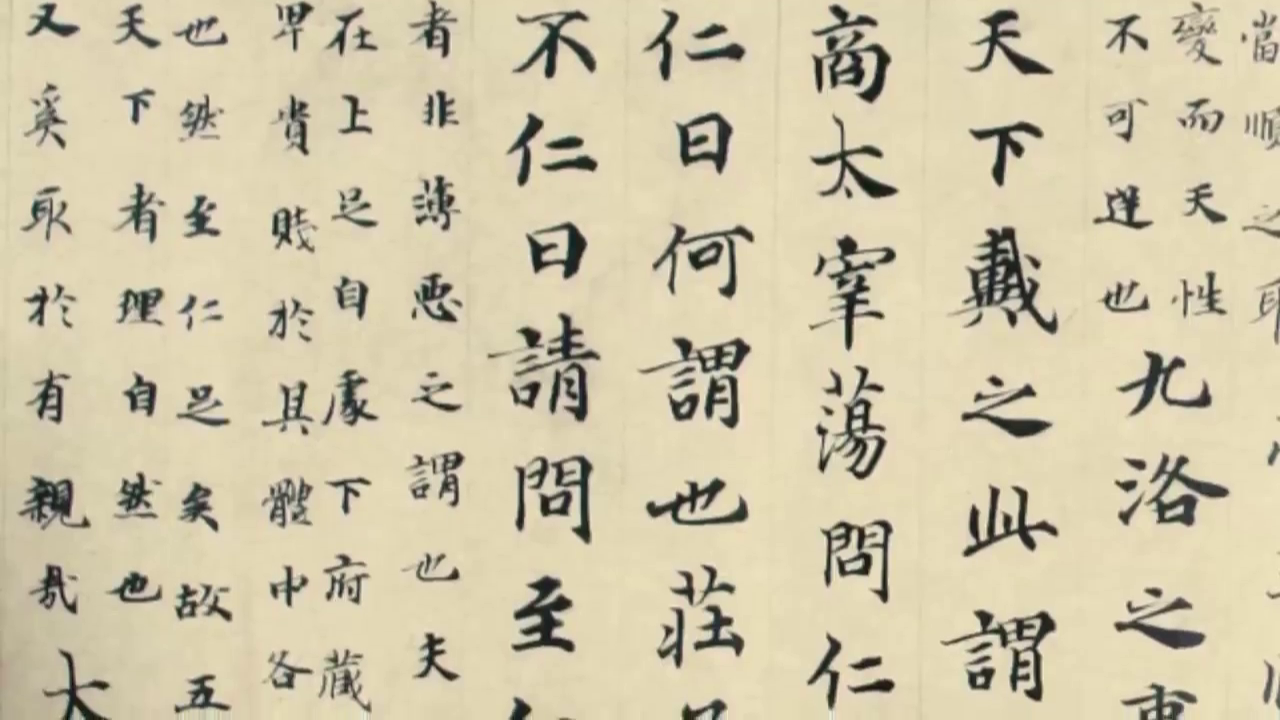} &
        \includegraphics[width=0.23\textwidth]{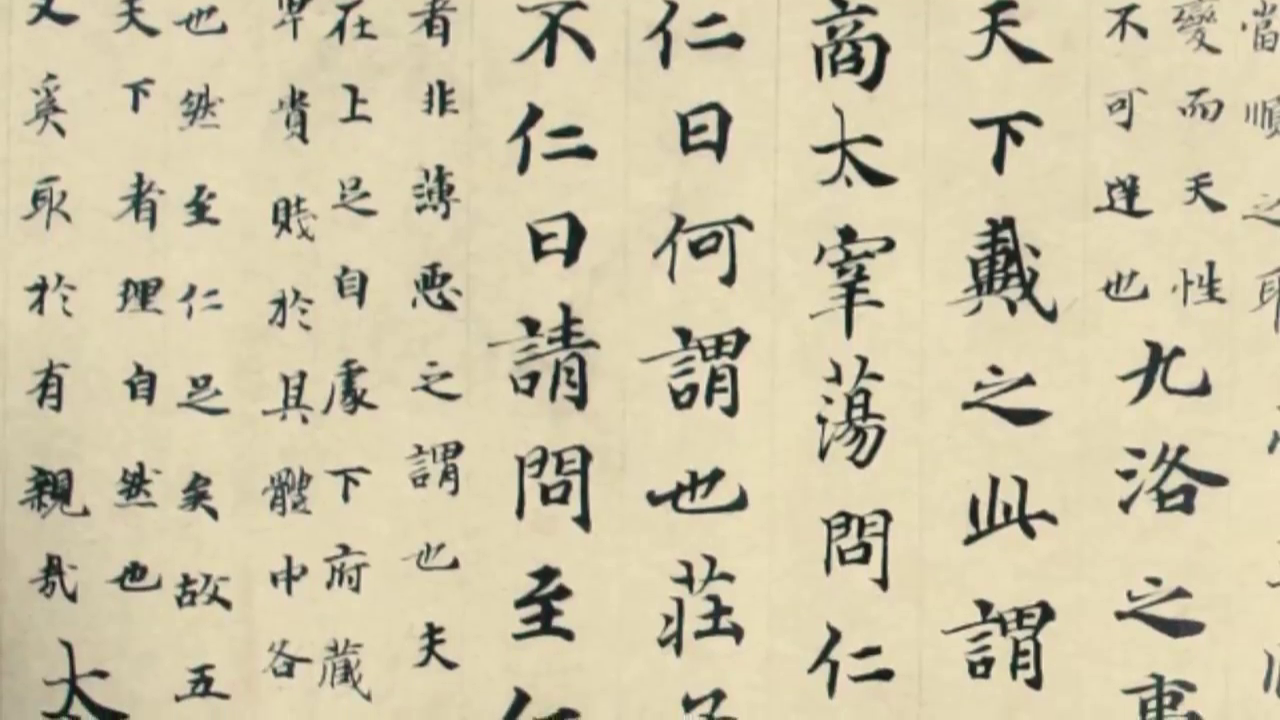} &
        \includegraphics[width=0.23\textwidth]{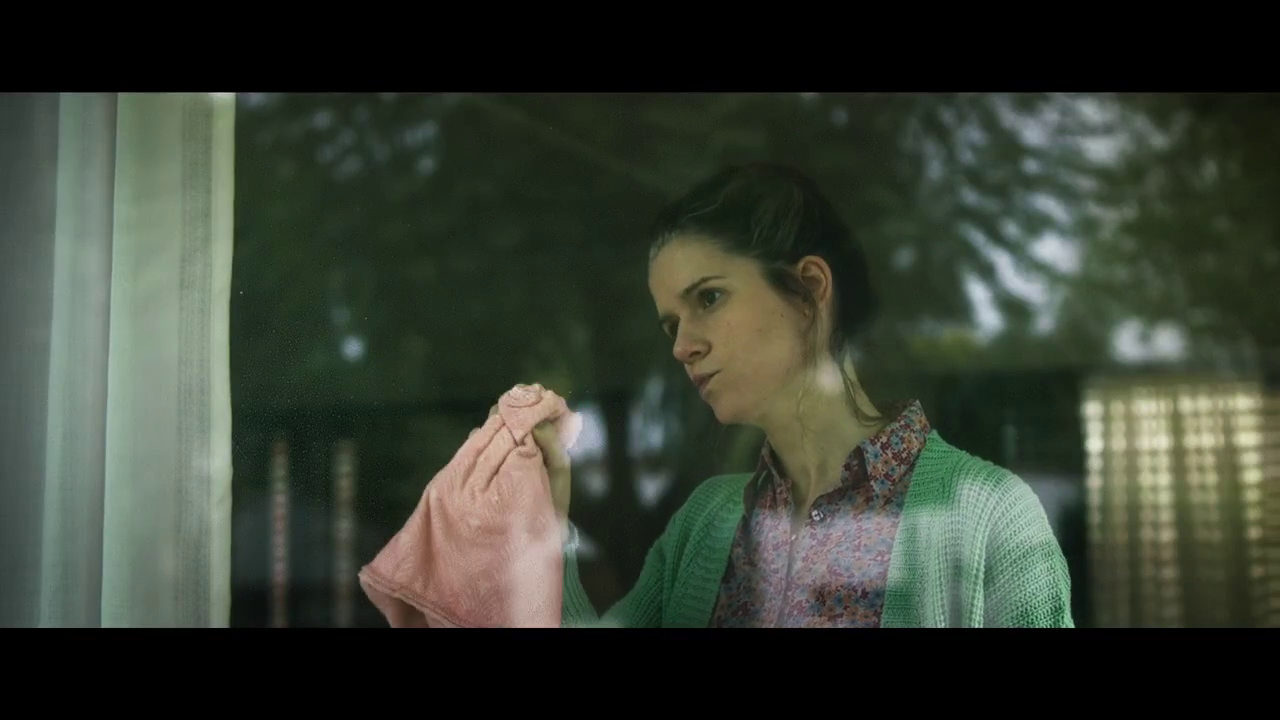} &
        \includegraphics[width=0.23\textwidth]{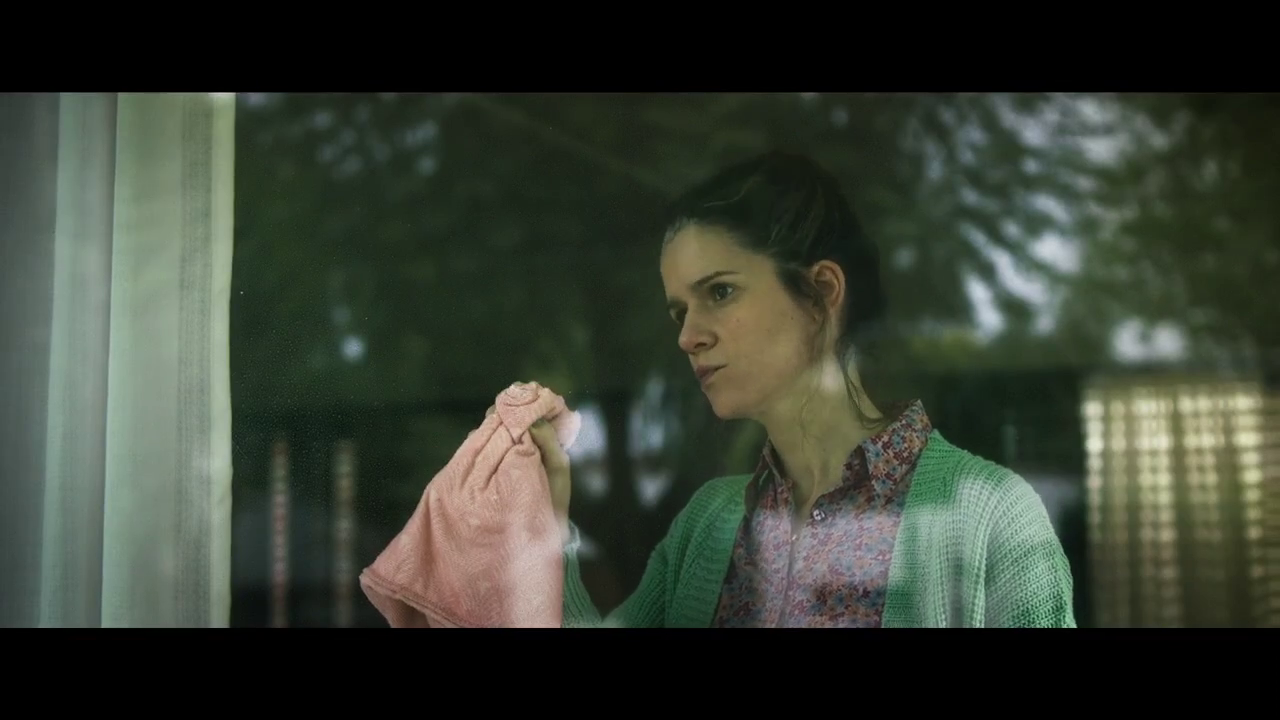} 
    \end{tabular}
    \caption{Clips of high (left) and low (right) OCR scores. }
  \label{fig:ocr_scores}
  % \vspace{-0.5cm}
\end{figure}

\textbf{Quality Statistics}
In addition to text detection, we considered image quality~\cite{huang2023vbench} and aesthetic scores~\cite{schuhmann2021laion400m} to enhance our analysis of trailer videos. These measures allowed us to evaluate frames' visual fidelity, clarity, and aesthetic appeal, providing more comprehensive insights for trailer analysis and editing. 
\begin{figure}[htb]
    \centering
    \begin{tabular}{cc|cc}
        \includegraphics[width=0.23\textwidth]{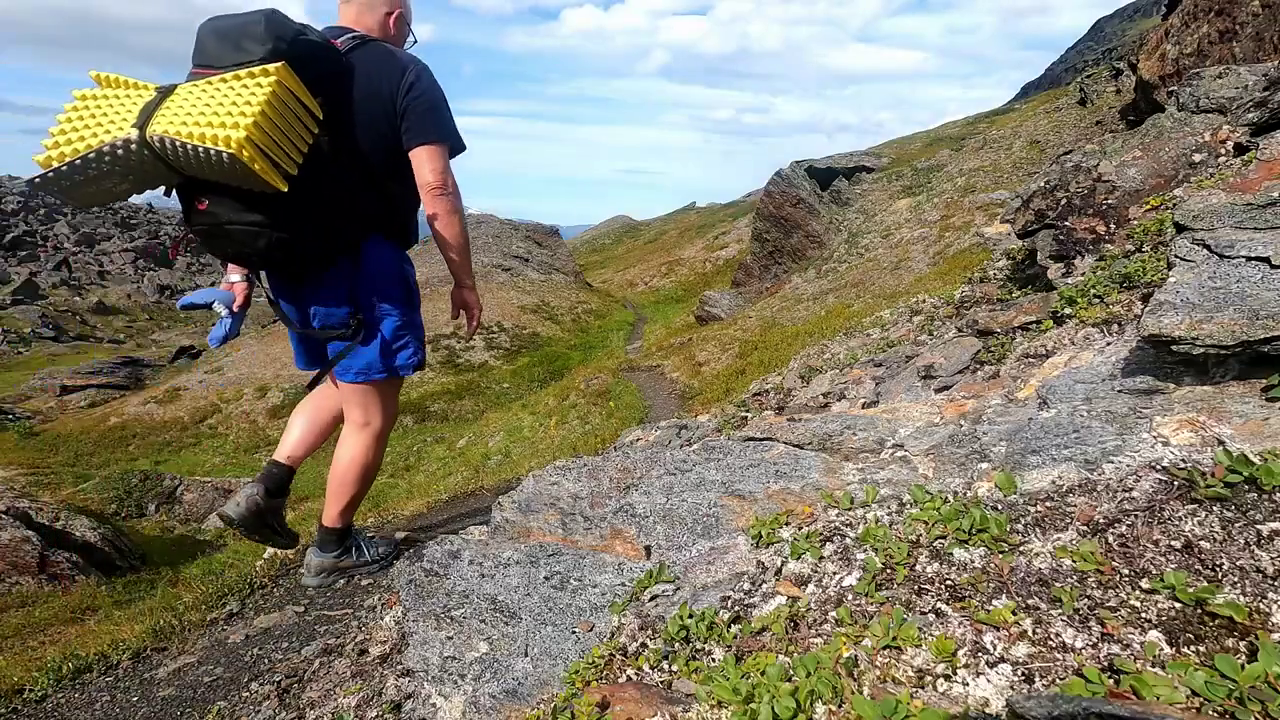} &
        \includegraphics[width=0.23\textwidth]{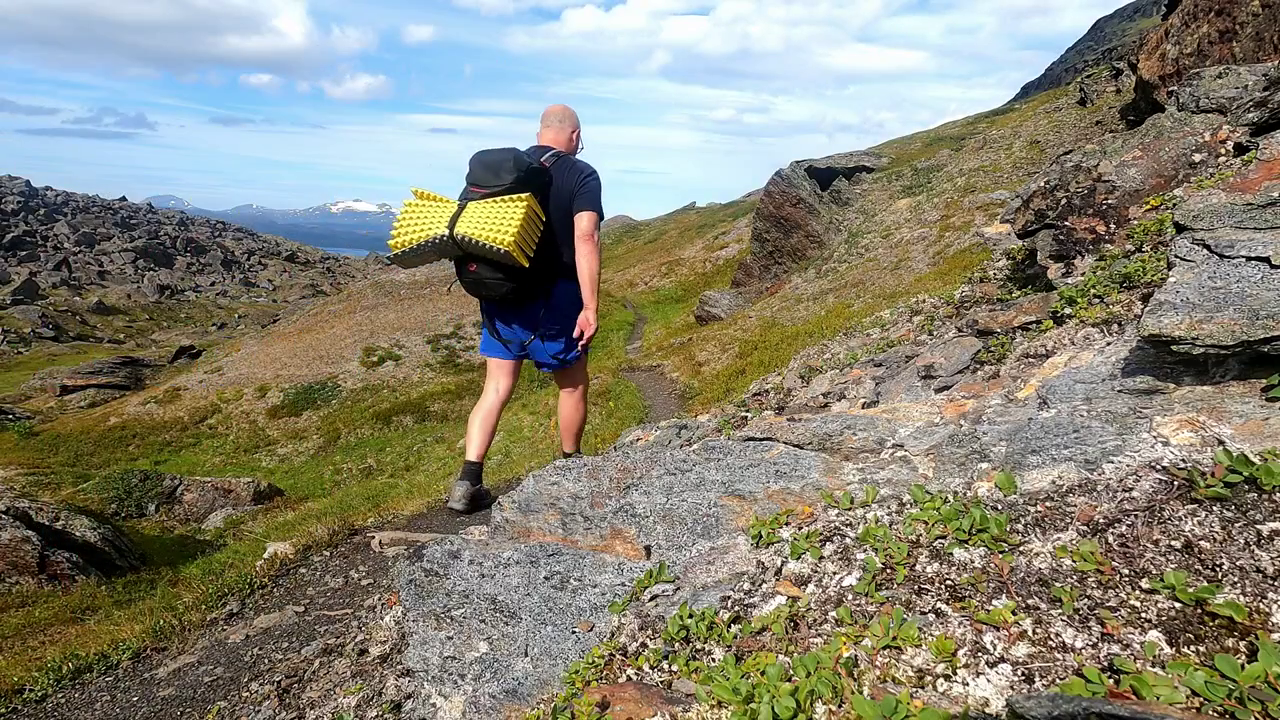} &
        \includegraphics[width=0.23\textwidth]{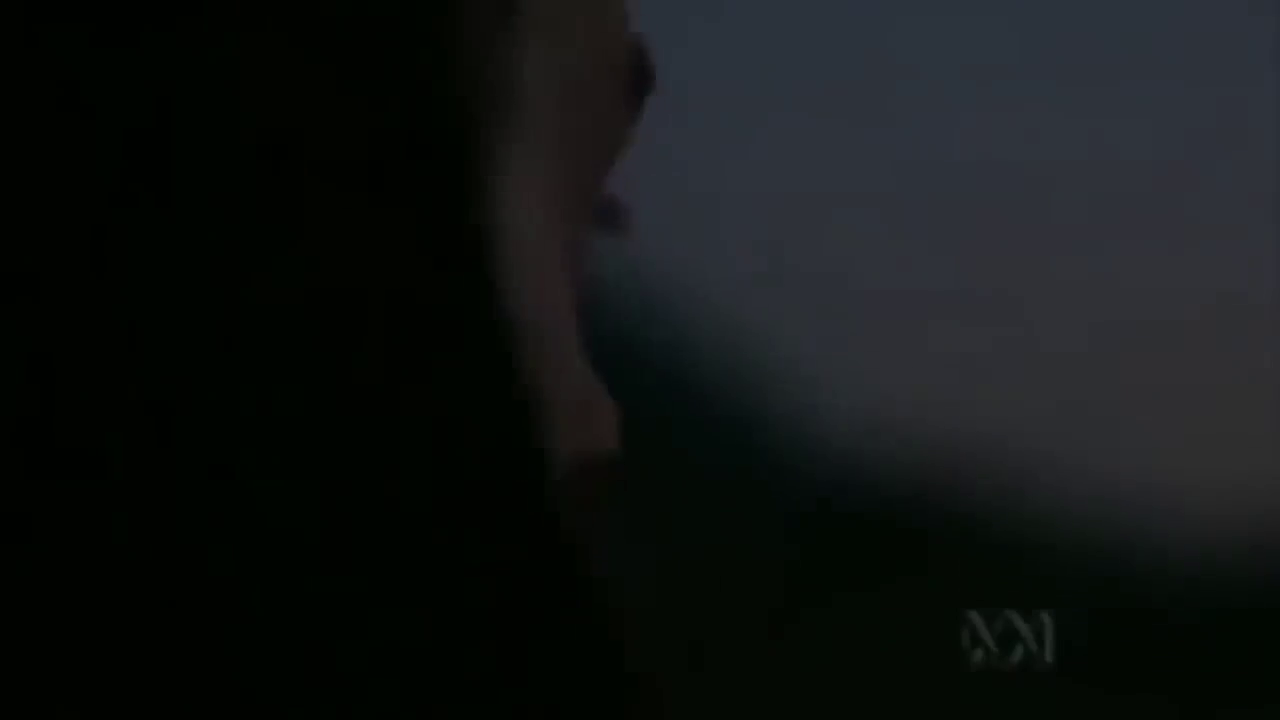} &
        \includegraphics[width=0.23\textwidth]{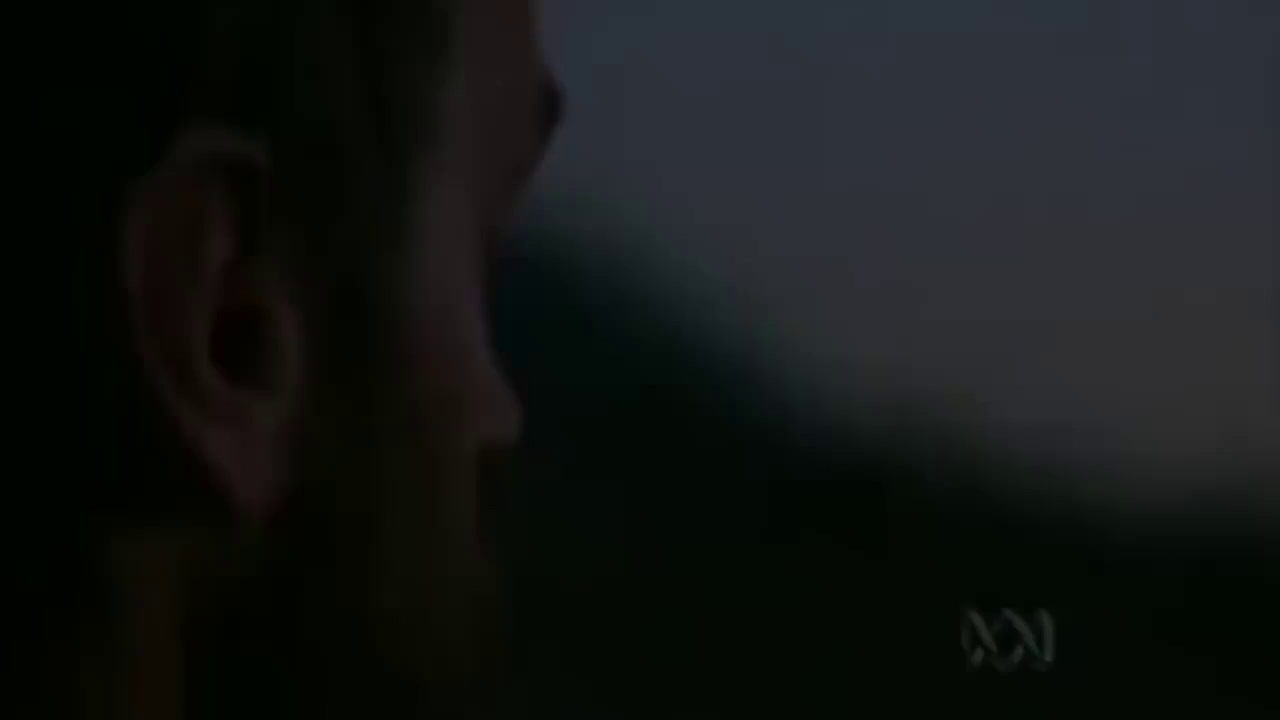} 
    \end{tabular}
    \caption{Clips of high (left) and low (right) image quality scores. }
  \label{fig:image_quality_scores}
  % \vspace{-0.5cm}
\end{figure}

\textbf{Audio Collection}
We extract the audio from video source segments with a sampling rate of 44.1 kHz. We use the PANNs~\cite{kong2020panns} algorithm to perform music event detection, and over 70\% of the audio segments contain music.

% Here Here Here ADD a paragraph about Music collection and quality

%%%%%%%%%%%%%%%%%%%%%%%%%%%%%%%%%%%%%%%%%%%%%%%%%%%%%%%%%%%%%%
%%%%%%%%%%%%%%%%%%%%%%%%%%captioning%%%%%%%%%%%%%%%%%%%%%%%%%%

\subsection{Video Captioning Pipeline}
\label{sec:method-ca}
The MMTrail contains many complex themes, like subtitles and character animations, as shown in ~\cref{fig:distribute}, which brings extraordinary complex work for video captioning. At the same time, smooth transition shots also make it impossible for traditional single-frame annotation methods to convey semantics coherently. Therefore, this section introduces a multi-temporal and multimodal caption pipeline containing a detailed video description from frame, motion, and music levels. 

\textbf{Frame Caption}
% 基本caption
\label{sec:method_fram_cap}
The auto-captioning pipeline has proven efficient in cutting-edge video generation foundation models. SVD~\cite{svd} and Pandas~\cite{panda70m} have given promising results and demonstrated the importance of high-quality frame captions for the generation model. We initially performed image-level captioning on the individual frames of the data. We employed coca~\cite{Yu2022CoCa} for each video clip to generate separate captions for three frames(first, middle, and last), resulting in relevant captions.

\textbf{Clip Caption}
Having obtained concise captions for three frames that capture the essential information, we aimed to obtain fine-grained captions and variations between frames in the video. We concatenated multiple frames into a comic strip format and employed the LLaVA~\cite{liu2024llavanext} image model to guide the description of the dynamic differences between frames. Additionally, leveraging a powerful multimodal language model, we incorporated OCR and more detailed summary descriptions to expand the information within the frame captions.

\textbf{Categories and Background}
Noticing the LLM-based caption has hallucinations when describing the frame, we further generate word-level labels to enhance the annotation of the main objects and background. Initially, we utilized LLaVA's QA capabilities to have the model answer questions about the background. Subsequently, through QA, we prompted the model to provide relevant category information. We conducted the word cloud in ~\cref{fig:word_cloud}. and certified caption quality by subjective experience in ~\cref{sec:exp}.

\begin{figure}[t]
  \centering
    \includegraphics[width=1\linewidth]{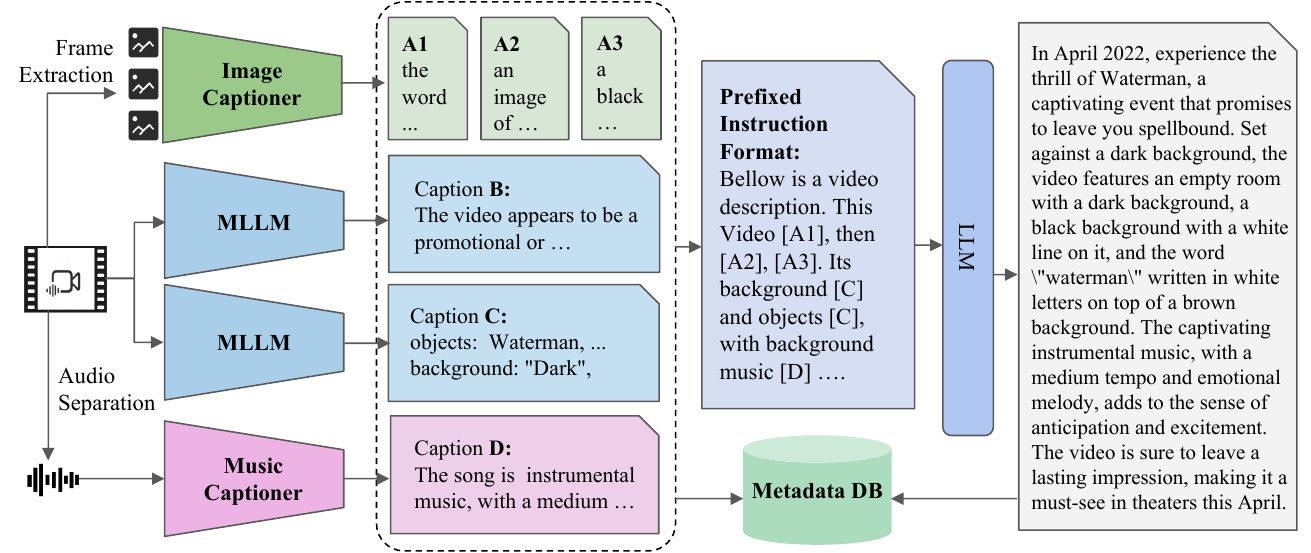}
  \caption{Data captioning pipeline. Starting from video clips, we extract frames and audio and then perform multiple rounds of captioning. A predefined instruction format combines multimodal captions, which serve as prompts for the language model and generate the final merged prompts.}
      \label{fig:caption_pipline}
      \vspace{-1em}
\end{figure}

\textbf{Music Caption}
Moreover, given that trailer music usually has a well-designed audio effect and background music, we applied the music caption on our dataset rather than a standard audio caption. In our work, we used MusicCaps~\cite{musiccap}, an LLM-based music captioning model. The caption format is well designed with its description pipeline, which first describes its sound quality, a generated speech style, and a detailed description of its instrument and music style. More examples are shown in the Appendix. We further evaluate the generation tasks and the text-to-music generation based on the music caption, which shows the efficiency of our captioning and dataset.

\textbf{OCR} The trailer contains many text animations. Captioning the context inside the movie can also be a challenging task. In this task, we utilize the OCR ability by LlaVA~\cite{liu2024llavanext} and caption the context for 5 frames of each video. We merge reliable text animation captions based on the previous method.

\textbf{General caption} Combining all the captions mentioned above, we use the language model llama2-13B~\cite{touvron2023llama2} to merge all the captions and generate complete and high-quality multimodal captions. We evaluate the caption accuracy and quality by human preference in ~\cref{sec:exp}.

\subsection{SubSet Separation}
\label{sec:method-subset}

We applied the frame caption~\cref{sec:method_fram_cap} for the full MMTrailer-20M video clips with 20M+ clips. We included the scale comparison with other large-scale datasets in ~\cref{tab:videodataset}, showing that our dataset is a large-scale video-language dataset. MMTrail has a resolution no smaller than 720p, and MMTrail-20M clips are 4.6s long on average.

\textbf{High-quality Subset}
, named MMTrail-2M, contains a detailed multimodal caption. Compared to the original distribution, we samples MMTrial-2M to create a high-quality subset using the following criteria:
1. We filter out clips with motion scores below 0.45 or above 50.
2. We only retain the clips within the top 85\% of image quality scores.
3. We only keep the clips within the top 85\% of aesthetic scores.
Furthermore, all clips in MMTrail-2M are longer than 4s and provided with all the captions, as shown in ~\cref{fig:caption_pipline}, including categories, background, frame captions, music captions, merged captions, etc.
We also compare MMTrail-2M with Video-Audio datasets as shown in ~\cref{tab:music}, MMTrail has a larger scale than existing datasets (Audioset~\cite{gemmeke2017audio} and Vggsound~\cite{chen2020vggsound}). 
 
\textbf{High-quality test set}
is extracted from the MMTrail-2M, we extract a fine-branded testing set that contains 1k video clips and multiple multimodal captions. Then, we manually adjust the merged caption to a manual caption to build a testing subset with trust-wise multimodal prompts. We test several tasks and models on the test set in ~\cref{sec:exp} to show the complexity and difficulties of MMTrail. The test set has 98.2 words of caption on average and includes 3.2hr video clips.
\begin{table}
  \caption{Comparison of MMTrail-2M and other Video-Audio Generation Dataset. For each dataset, we list the following information in each column: dataset name (Dataset), public year (Year), average duration per clip (Dur./Clip), total number of clips ($\#$Clips), total number of hours ($\#$Hours).}
  \label{tab:music}
  \centering
      \setlength\tabcolsep{20pt}%调列距
        \resizebox{0.95\columnwidth}{!}{
  \begin{tabular}{lcccc}
  \toprule
    Dataset     & Year & Dur./Clip & \#Clips & \#Hours   \\
    \midrule
    % Trailer-Audio  & 2024  & \tzy{xiaowe} & \tzy{xiaowe} & \tzy{xiaowe}  \\
    Audioset~\cite{gemmeke2017audio} & 2017  & 10s & 2M & 5.8khr    \\
    Vggsound~\cite{chen2020vggsound} & 2020  & 10s & 210K  & 550hr   \\
    MMTrail-2M     &  2024  & 13.8s & 2M & 8.2khr  \\
    MMTrail-Test     &  2024 & 11.6s & 1k &  3.2hr  \\    
    \bottomrule
  \end{tabular}
  }
      \vspace{-1em}
\end{table}

% \vspace{-0.5em}
\begin{figure}[t]
  \centering
    \includegraphics[width=\linewidth ]{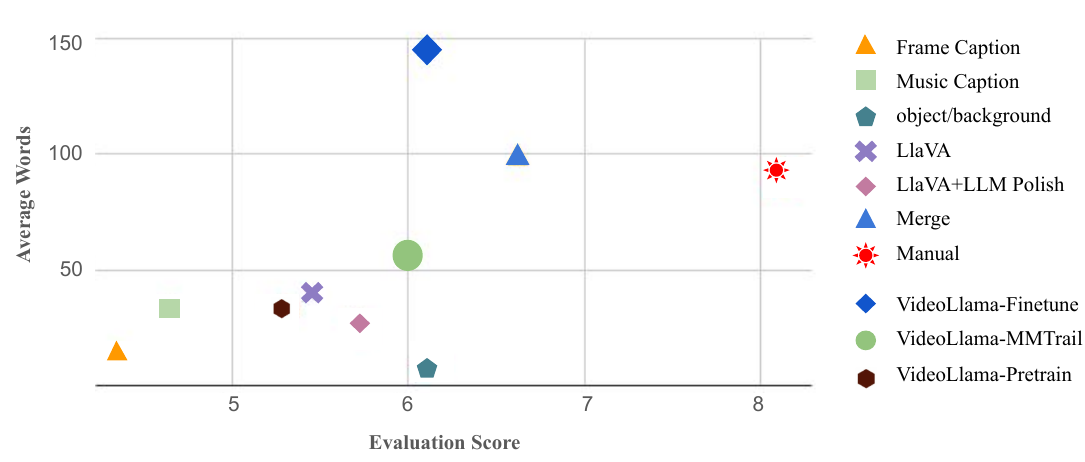}
  \caption{Human evaluation results of the captioning models on the MMTrail-Test. The X-axis is the average evaluation score from 0-10, and the Y-axis is the average word numbers.}
      \label{fig:human_eval}
        \vspace{-5mm}
\end{figure}

\vspace{-1em}
\section{Experiments}
\label{sec:exp}

This section presents comprehensive experiments on multiple tasks to demonstrate our dataset's effectiveness, diversity, complexity, and difficulty. 

\subsection{Multimodal Captioning}
\label{sec:exp_cap}
% TO be checked

We present the results of our human evaluation of video caption quality in ~\cref{fig:human_eval}. Ten videos were randomly selected from the MMTrail-Test dataset. They were rated on a scale of 0 to 10 based on general impressions, including aspects such as correctness, level of detail, richness, and fluency. The results of more than 100 sets of samples indicate that the manually adjusted prompts rating of 8.12 outperforms the auto-caption pipeline, while our merged captions achieve the second-best performance of 6.62. Despite being short and straightforward, object/background labels achieve a 6.02 evaluation score, demonstrating more correctness than other captions. Frame caption, music caption, and llaVA caption obtain 4.3, 4.6, and 5.4, respectively, and these findings demonstrate the effectiveness of our captions and highlight the quality of our labeled captions by human annotators.

\subsection{Video Generation}
\label{sec:exp_gen}

\begin{table}[t]
\centering
\caption{Comparison of VideoCrafter-2.0 and VideoCrafter-2.0(MMTrail) on 9 different dimensions. For every dimension, a higher score is better.}
\setlength\tabcolsep{18pt}%调列距
\label{tab:videocrafter_evaluation}
    \resizebox{0.95\columnwidth}{!}{
    \begin{tabular}{l|c|c}
  \toprule
    \textbf{Dimentions($\uparrow$)} & VideoCrafter-2.0 & \textbf{VideoCrafter-2.0(MMTrail)} \\
    \midrule
    temporal style & 25.84 & 24.61 \\
    appearance style & 25.13 & 24.10 \\
    image quality & 67.22 & 69.78 \\
    dynamic degree & 42.50 & 43.50 \\
    motion smoothness & 97.73 & 98.33 \\
    temporal flickering & 98.41 & 98.50 \\
    Subject consistency & 96.85 & 98.62 \\
    background consistency & 98.22 & 98.40 \\ 
    Overall consistency & 28.23 & 25.33 \\ \midrule
    \textbf{Sum} & 64.45 & \textbf{64.57} \\  
    \bottomrule
    \end{tabular}
    }
    \vspace{-0.5em}
\end{table}

\vspace{-1.0em}
\begin{figure}[t]
  \centering
    \includegraphics[width=\linewidth ]{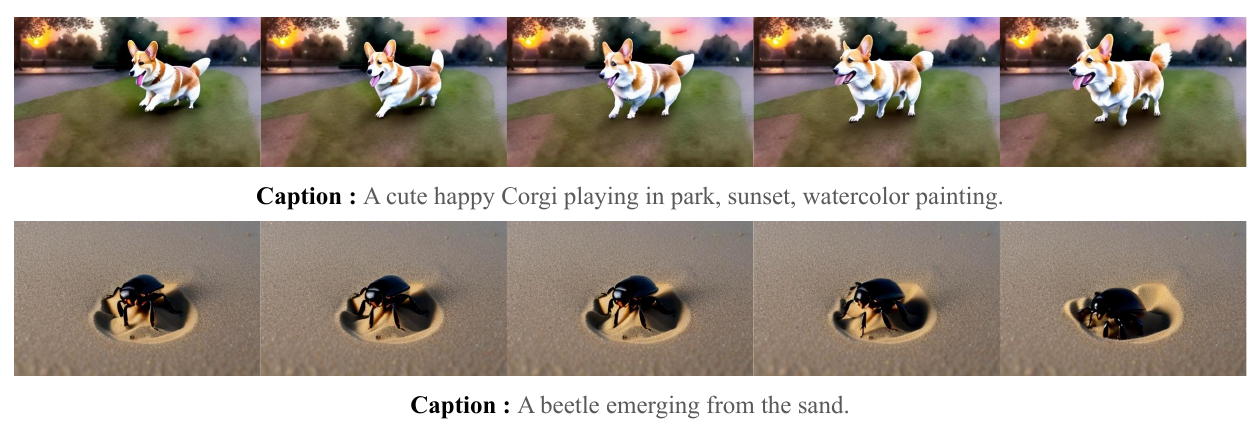}
  \caption{Two generation result of Videocrafter(MMTrail). The caption is from the VBench~\cite{huang2023vbench} evaluation prompts list; the given example shows the high quality in motion and object consistency. }
      \label{fig:gen_example}
    \vspace{-1em}
\end{figure}

The model was fine-tuned on the VideoCrafter-2.0~\cite{long2024videodrafter} dataset using 8 Tesla-H800 GPUs with a batch size of 3 for 10,000 steps at a learning rate of 6e-6. The training data was randomly sampled from the MMTrail-2M dataset, using the video captions as input.
The evaluation results, shown in ~\cref{tab:videocrafter_evaluation}, include 9 matrices on the VBench~\cite{huang2023vbench}, indicating that fine-tuning the model on the MMTrail-2M dataset led to improvements of 0.6 in motion smoothness and 1.77 in subject consistency, with a slight overall performance boost(0.12 higher) compared to the official VideoCrafter-2.0 checkpoint.
Visual examples of the generated content are provided in ~\cref{fig:gen_example}, and additional demonstrations and experiment details will be included in the supplementary material.
This thorough evaluation and comparison of the tuned model's performance on critical metrics provides valuable insights into the effectiveness of the fine-tuning process and the potential benefits of leveraging the MMTrail-2M dataset for video generation tasks.

\subsection{Video Understanding}
\label{sec:exp_understand}
\begin{table}[t]
  \caption{ Comparison of Video-LlaMA model performance on the Trailer-Test dataset. The figure shows the results of three different versions of the Video-LlaMA model across five evaluation metrics, and the Video-LlaMA(MMTrail) version performs better on most evaluation indicators.}
  \setlength\tabcolsep{11pt}%调列距
  \label{tab:video_mllm}
  \centering
      \resizebox{0.95\columnwidth}{!}{
  \begin{tabular}{lccccc}
  \toprule
    Model & BLEU-4$\uparrow$     & M$\uparrow$ & ROGUE-L$\uparrow$ & CIDEr$\uparrow$ &BERT$\uparrow$ \\
    \midrule
    Video-LlaMA(Pretrain) &0.52&4.57&11.57&0.09&84.42\\
    \midrule
    Video-LlaMA(Finetune)  &3.94&14.05&22.67&2.45&85.48\\
    \midrule
    Video-LlaMA(MMTrail) &5.59&13.83&24.97&24.79&87.21 \\
    \bottomrule
  \end{tabular}
  }
\end{table}
\textbf{Experiment Setting}
To evaluate the capability of our dataset in multimodal video understanding, we choose Video-LLaMA~\cite{videollama} as the baseline for the video captioning task. We use same model and training config as Video-LlaMA, which useVicuna-v0-7B as llama model~\cite{zheng2023judging}, ViT~\cite{dosovitskiy2021image} and Q-Former ~\cite{zhang2023vision} as the video encoder and the linear projection layer from MiniGPT-4~\cite{zhu2023minigpt}. We train 4 epochs by MMTrail-2M, each containing 2500 iters with batch size 32. We compare it with two official model weights: the pre-train Video-LlaMA weight on WebVid (2.5M video-caption pairs) and the fine-tuned Video-LlaMA.

\textbf{Evaluation Metric}
We evaluate video understanding models on the MMTrail-Test. As for the evaluation metric, we choose the commonly used metrics in text generation tasks-BLEU-4~\cite{papineni-etal-2002-bleu}, ROGUE-L~\cite{lin-och-2004-automatic}, METEOR~\cite{banerjee-lavie-2005-meteor}, and CIDEr~\cite{vedantam2015cider} to evaluate our result.
All the metrics are computed using the pycocoevalcap~\cite{lin2015microsoft} package. We also use BERTScore~\cite{zhang2020bertscore} to evaluate the contextual similarity for each token in the ground truth and the predicted captions. The results are reported in ~\cref{tab:video_mllm}. The official weights show relatively low performance, highlighting the challenge of MMTrail, and the data distribution differs from their training data.

In addition, we also evaluated three checkpoints from Video-LlaMA~\cite{videollama} by human evaluation in ~\cref{fig:human_eval} and found that the Video-LlaMA-MMTrail evaluation result slightly lags behind Video-LlaMA-Finetune but performs significantly better than Video-LlaMA-Pretrain. We provide further details in ~\cref{sec:exp_understand} for a more comprehensive understanding of our model.

\subsection{Music Generation}
\label{sec:exp_v2m}
We used text-to-music generation to evaluate the effectiveness of the video-music pair data and the labeled video caption and music caption. 
We use MusicGen~\cite{copet2024simple} to generate music based on our video caption (VideoCap2Music) and music caption (MusicCap2Music). We use Kullback-Leibler Divergence (KL), Inception score (ISc), Frechet distance (FD), and Frechet Audio Distance (FAD)~\cite{kilgour2018fr} to evaluate the generated music. Besides, we use the ImageBind-AV score (IB) to evaluate the audio-visual alignment between the video and the generated music. For the model with text input in Tab.~\ref{tab: music_gen}, compared with video caption, the evaluation results on music caption are 0.13 better in KL, 0.65 in ISc, 3.64 in FD, and 1.21 in FAD, showing the domain gap of multimodal descriptions.
This comparison shows that there is still a significant research gap between caption-music-video.

\begin{table}[t]
  \vspace{-0.5em}
  \caption{Music generation evaluation results on the MMTrail-Test. We compare two types of captions and their 5 metrics. The results show that music caption performs better in music generation tasks.}
  \label{tab: music_gen}
  \centering
    \setlength\tabcolsep{15pt}%调列距
        \resizebox{0.95\columnwidth}{!}{
  \begin{tabular}{lcccccc}
  \toprule
    Method  & Input & KL$\downarrow$ & ISc$\uparrow$ & FD$\downarrow$ & FAD$\downarrow$ & IB$\uparrow$ \\
    \midrule
    VideoCap2Music & Text  & 3.22 & 1.79 & 57.17 & 15.04 & 0.09     \\
    MusicCap2Music & Text & 3.10 & 2.44 & 53.53 & 13.83 & -    \\
    % \midrule
    % Video2Music & Video & 0.988 & 1.231 & 48.144 & 5.078 & 0.183     \\
    \bottomrule
  \end{tabular}
  }
  \vspace{-0.5em}
\end{table}

\section{Conclusion}
\label{sec:conclu}
We introduce MMTrail, a comprehensive and accurate multi-modality visual-audio dataset to address the dataset gap. By utilizing the inherent value of trailers, which integrate visual, audio, and contextual elements, MMTrail offers detailed and precise multi-modality annotations. Our systematic captioning framework adaptively merges visual and musical perspectives, ensuring that the annotations capture the richness of multimodal content. Experimental results demonstrate the high quality of the MMTrail dataset, its effectiveness for fine-grained multimodal-language model training, and a variety of downstream applications. We believe this innovative dataset will unlock new possibilities in video content generation and significantly advance research in visual-audio understanding. The comprehensive and diverse nature of MMTrail makes it a valuable asset for the research community, paving the way for novel applications that leverage the power of multimodal learning.
\appendix

\section{Captioning Pipeline}
\label{sec:sub_cap}
\subsection{Music Caption}
% Trailer audio 很杂乱，包含各种轨道。
Methods like Chatmusician\cite{yuan2024chatmusician} provide captions for music, however, as trailer videos often contain a cacophony of audio tracks, which typically include background music and vocals, captioning audio poses a significant challenge. Therefore, to achieve optimal music captioning results, we initially utilize Demucs~\cite{demucs} for vocal separation on each audio clip. Subsequently, LP-MusicCap~\cite{doh2023lp} is leveraged to caption the resultant audio, devoid of vocals, from the separation process. 

\subsection{Speech Recognition}

\begin{figure}[t]
  \centering
  % \vspace{-6mm}
  \begin{subfigure}{1\linewidth}
      \includegraphics[width=\linewidth]{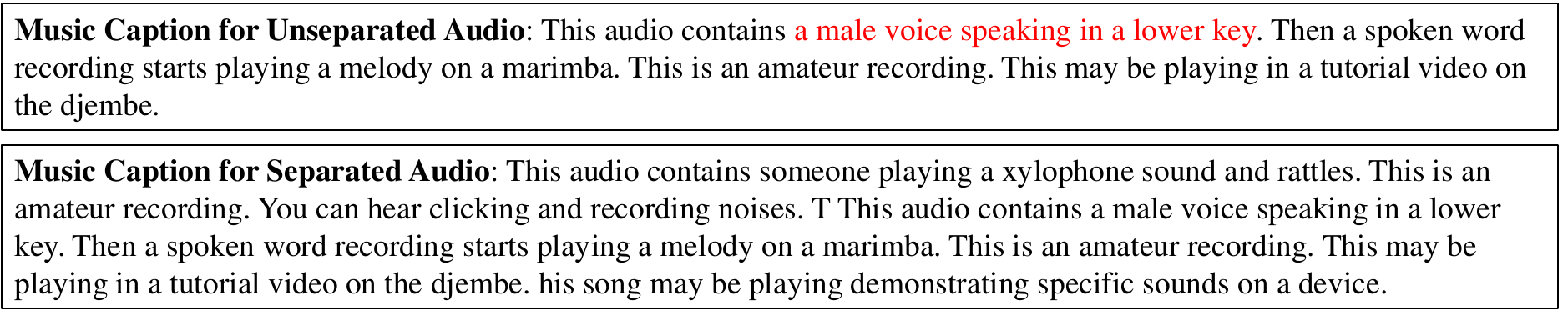}      
      \label{fig:mmerge_example_1}
      \caption{}
  \end{subfigure}
  \begin{subfigure}{1\linewidth}
      \includegraphics[width=\linewidth]{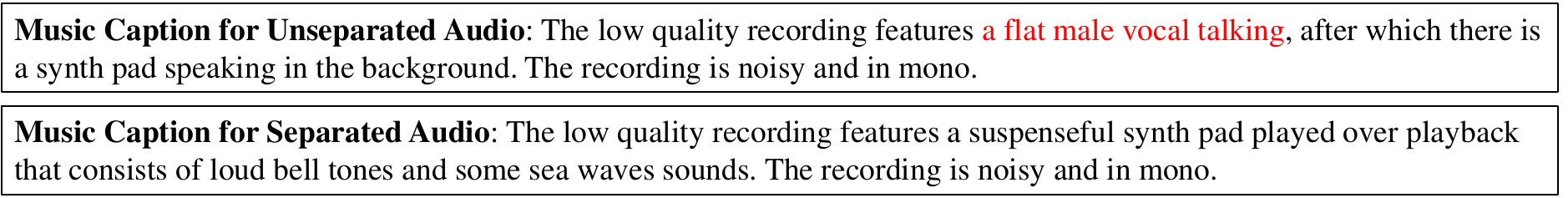}       
      \label{fig:mmerge_example_2}
      \caption{}
  \end{subfigure}
  \begin{subfigure}{1\linewidth}
      \includegraphics[width=\linewidth]{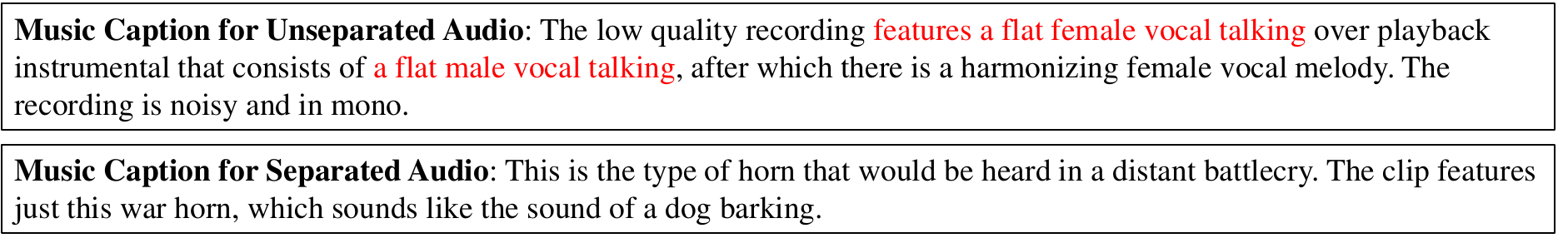}     
        \label{fig:mmerge_example_3}
      \caption{}
  \end{subfigure}
  \begin{subfigure}{1\linewidth}
      \includegraphics[width=\linewidth]{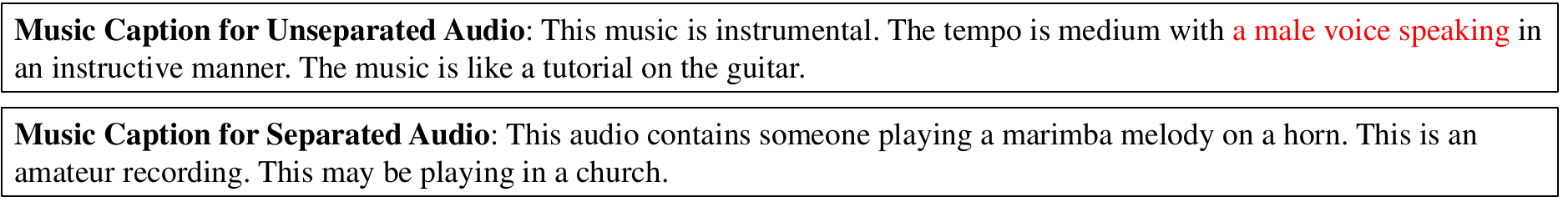}     
        \label{fig:mmerge_example_4}
      \caption{}
  \end{subfigure}
    \caption{We provided an extra comparison of the music caption before and after the track separation. With our separation, the caption includes the description of the human voice as highlighted in red.}
  \label{fig:music_example}
  % \vspace{-0.5cm}
\end{figure}
Besides, the speech is also an important track of the trailer audio. Therefore, we further turn the speech into the text.
Similar to our approach in music captioning, we use Demucs initially to perform vocal separation on each audio clip. Following this, Whisper is utilized to caption the separated vocal audio.

% \begin{enumerate}
% \subsection{Frame Caption}
% We utilize LLava-v1.6-vicuna-7B as the image understanding model to generate captions for video clips. Initially, we filter out clips with durations of less than 1.0 seconds. Subsequently, we sample three frames at the fractions of 0.2, 0.5, and 0.8 of each clip's total duration. The prompt is as follows:
%  \begin{lstlisting}
%  Please describe the image in detail.
%  \end{lstlisting}

% \clearpage

\begin{figure}[t]
    \centering
    \includegraphics[width=0.8\linewidth ]{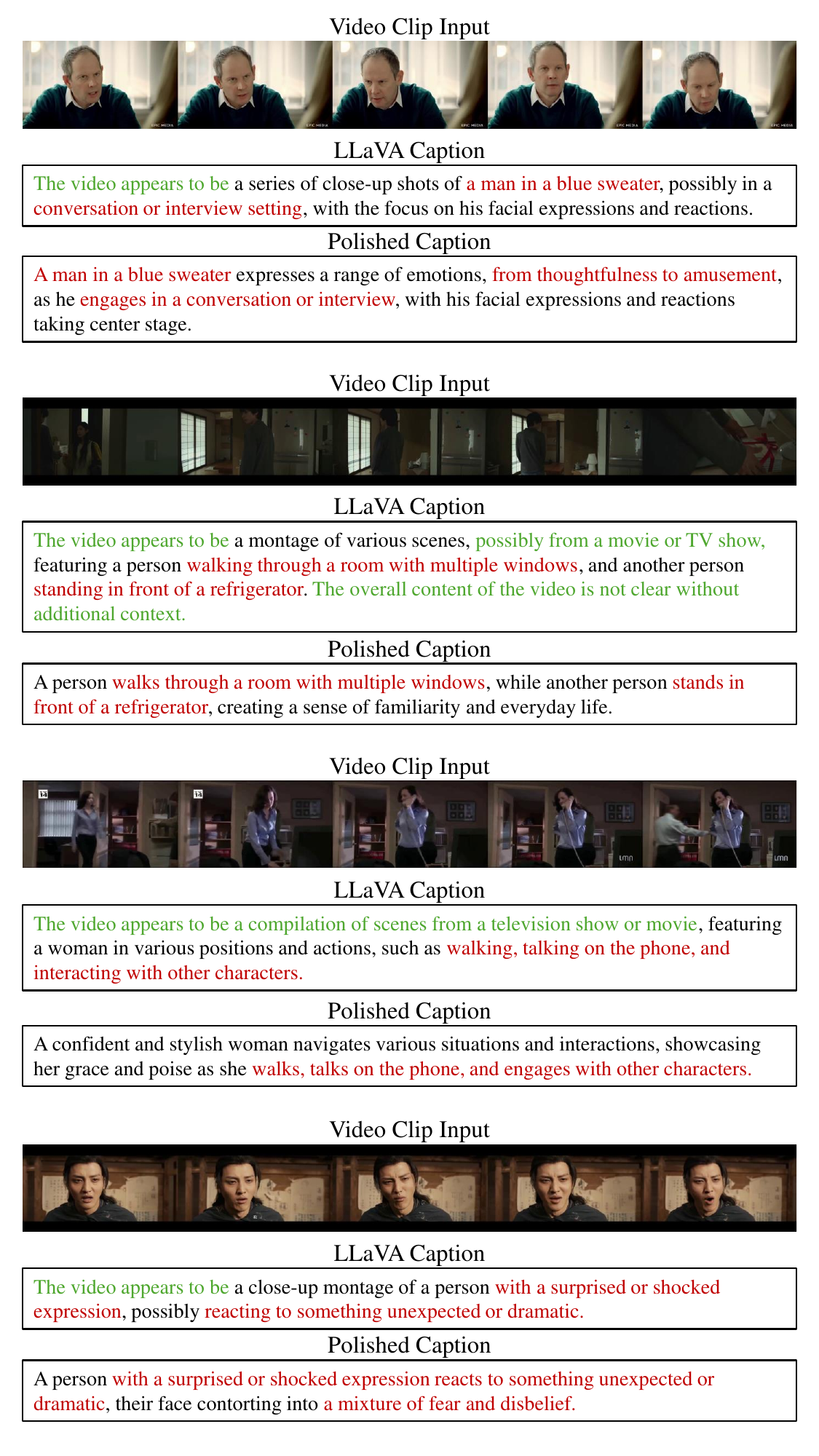}
    \caption{While using LLaVA as the backbone model for our video caption, we find that it contains a lot of "Oral habit", as highlighted in green text in this figure. We further apply a language model to reception the sentence. }
    % from an in-the-wild trailer video that contains multi-modality annotation and evaluation.}
    \label{fig:cap_example}
\end{figure}
\subsection{LLaVA Caption}
We use the image understanding model LLaVA-13b~\cite{liu2024llavanext} to caption each of our video clips. Specifically, for each clip, we sample frames at positions 0.1, 0.3, 0.5, 0.7, and 0.9, and then horizontally concatenate them into a single image, which serves as input for LLaVA-13b to generate captions. We construct the caption prompt as follows:
    \begin{lstlisting}
    These are some keyframes of a video. 
    Please use one sentence to summarize the content of the video in detail. 
    Summarize the content of the entire video but not describe keyframes frame by frame.
    \end{lstlisting}

\subsection{Polish LLaVA Caption}
Moreover, we observe that the results of LLaVA captioning often contained some redundant information, as illustrated by the green sections in Figure~\ref{fig:cap_example}. Therefore, we use LLaMA-13b to refine the LLaVA captions, eliminating much of the extraneous content and rendering the final captions more in line with human expression. We construct the prompt as follows:
   \begin{lstlisting}
    [{caption}] This is a description of a video. 
    Please polish it to an overall video description in one sentence and give me only the content of the video. 
    Do not use the words 'frame' and 'video'. 
    Describe the content of the video directly, which means do not start with 'The video...' or something like that.
    Do not add extra information that is not included in the original description.
    Here is an example: A dancer in a vibrant orange skirt and gray jacket moves gracefully across the stage, her movements fluid and expressive. 
    \end{lstlisting}
% 总结一下 最后得到了好caption
By combining frames and secondary polishing, we finally obtain high-quality captions that contain the main content information of the clips.

\subsection{Merge Caption}
\begin{figure}[t]
  \centering
  % \vspace{-6mm}
  \begin{subfigure}{1\linewidth}
      \includegraphics[width=\linewidth]{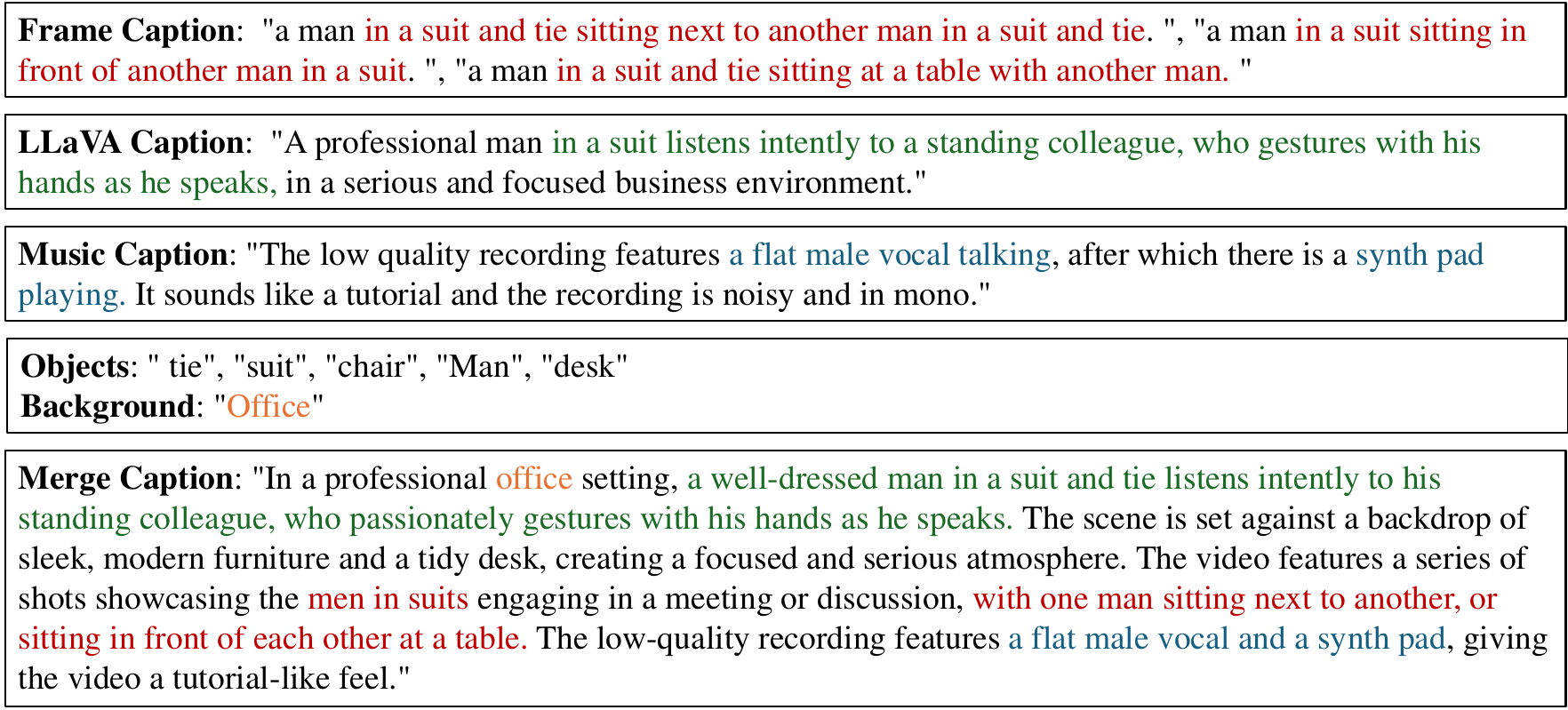}      
      \label{fig:merge_example_1}
      \caption{}
  \end{subfigure}
  \begin{subfigure}{1\linewidth}
      \includegraphics[width=\linewidth]{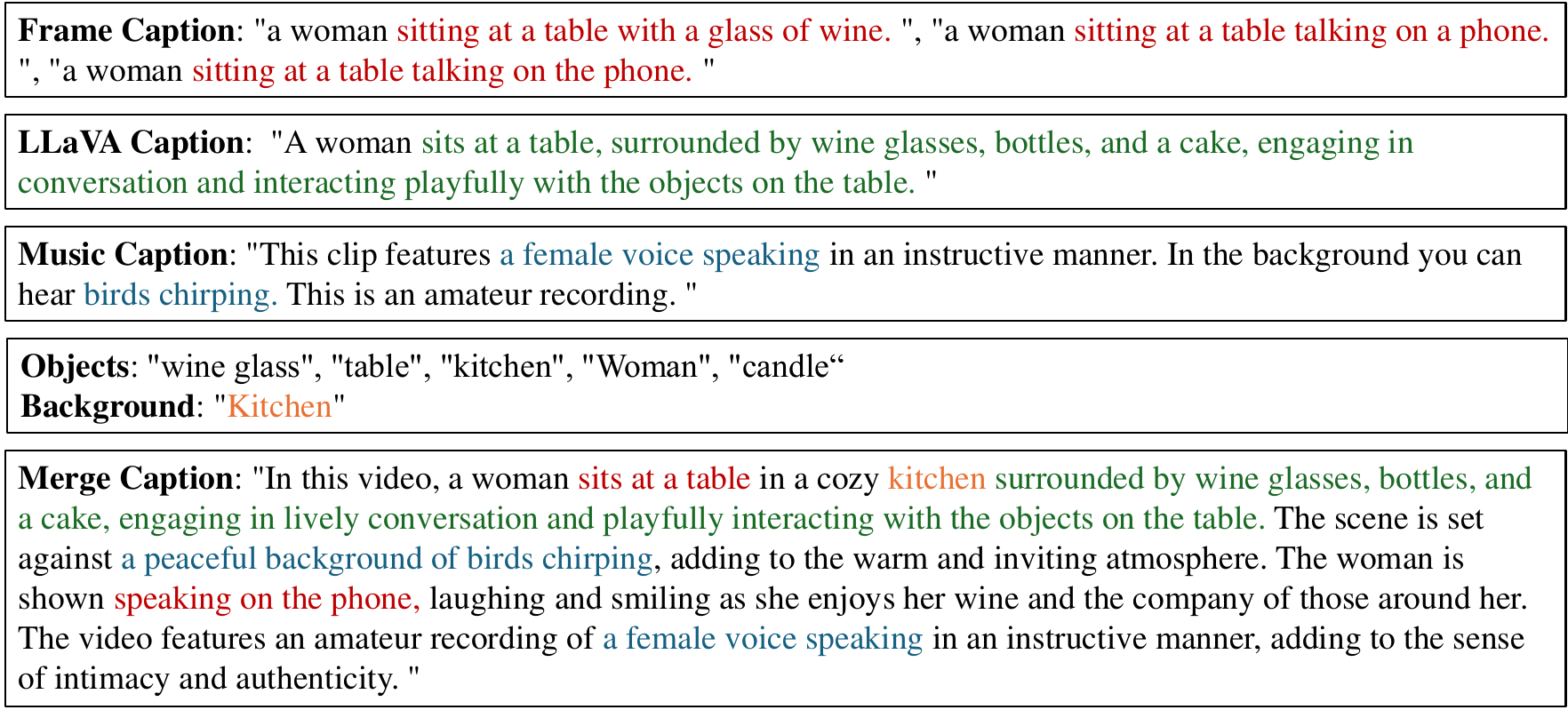}       
      \label{fig:merge_example_2}
      \caption{}
  \end{subfigure}
  \begin{subfigure}{1\linewidth}
      \includegraphics[width=\linewidth]{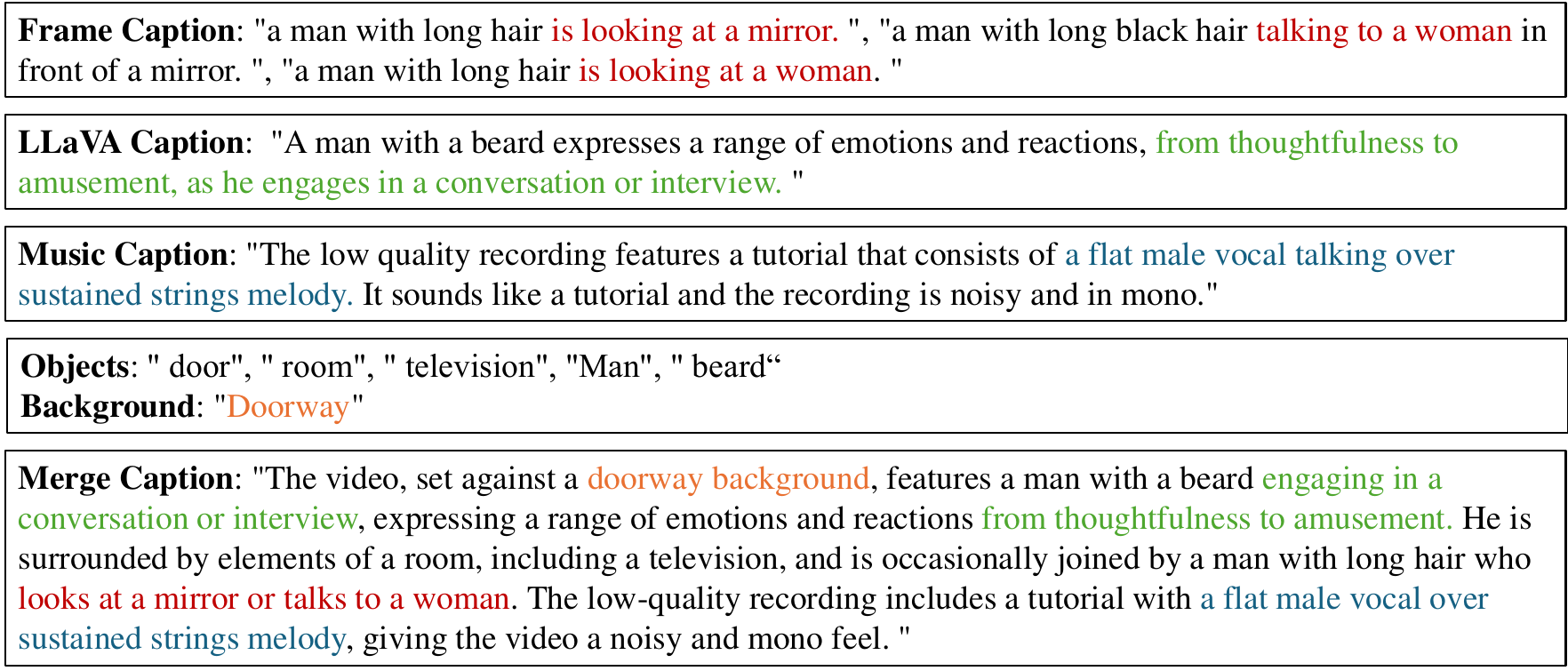}     
        \label{fig:merge_example_3}
      \caption{}
  \end{subfigure}
    \caption{We demonstrate more examples of the merged captions. As shown in the examples, all key information from different captions is merged together into a fluent paragraph. }
  \label{fig:merge_example}
  % \vspace{-0.5cm}
\end{figure}
% \section{Captioning Pipeline}
% Finally
We use LlaMA-13B to merge the multiple captions, by constricting a pre-designed caption prompt as follows:
    \begin{lstlisting}
    There are some descriptions of a video, 
    including video caption, music caption, background caption and the main objects in the video. 
    Please combine all the descriptions into an overall description of the video in only one paragraph.
    Finally, please rewrite and polish it into an overall video description in one paragraph and give me only the content of the video. 
        Background: {background}
        Main objects: {objects}
        Video caption 1: {image_cap}
        Video caption 2: {frame_cap}
        Music caption: {music_cap}
    \end{lstlisting}
% example, 分析一下结果
As illustrated in Figure~\ref{fig:merge_example}, we use the LLaMA-13B to merge frame captions, lava captions, music captions, objects, and background into a final merge caption. By merging multiple captions, the merge caption offers a comprehensive and detailed description of the audiovisual content of videos. Its components mutually complement one another, ensuring that every aspect of the narrative receives attention and providing unique insights into various facets of the video content.
% \clearpage

\section{Additional Experiments}
\label{sec:sub_exp}
\subsection{Video Understanding Model}
We use VideoLLaMA as our base model, which contains a llama model, a q-former, and a llama projection model. For LLM, we use llama vicuna-v0-7B.  
We also use model weight pretrained on minigpt4 to initialize our llama projection. In the training stage, we train from scratch and froze the Visual Transformer and LLM models. Only the visual q-former and the llama projection are trained in the pretraining stage. Following the video llama setting, we use a batch size of 32, and each video uniformly samples 8 frames with a 224 resolution. We train 5 epochs on our 2M training data. Each epoch contains 2500 iters with a linear warmup cosine lr. The weight decays are 0.05, and the warmup lr is 1e-6 with 2500 warmup steps, a 1e-4 init lr, and an 8e-5 minimum lr. The results can be seen in Table~\ref{tab:video_mllm+}
\begin{table}[h]
  \caption{ Comparison of Video-LLaMA model performance on the Trailer-Test dataset. The figure shows the results of three different versions of the Video-LLaMA model across five evaluation metrics, and the Video-LLaMA(MMTrail) version performs better on most evaluation indicators.}
  \setlength\tabcolsep{11pt}%调列距
  \label{tab:video_mllm+}
  \centering
      \resizebox{0.95\columnwidth}{!}{
  \begin{tabular}{lccccc}
  \toprule
    Model & BLEU-4$\uparrow$     & M$\uparrow$ & ROGUE-L$\uparrow$ & CIDEr$\uparrow$ &BERT$\uparrow$ \\
    \midrule
    Video-LLaMA(Pretrain) &0.52&4.57&11.57&0.09&84.42\\
    \midrule
    Video-LLaMA(Finetune)  &3.94&\textbf{14.05}&22.67&2.45&85.48\\
    \midrule
    Video-LLaVA~\cite{videollava}  &2.80&8.18&21.27&7.92&\textbf{87.61}\\
    \midrule
    Video-LLaMA(MMTrail) &\textbf{5.59}&13.83&\textbf{24.97}&\textbf{24.79}&87.21 \\
    \bottomrule
  \end{tabular}
  }
\end{table}

% \subsection{VideoCrafter-2 Finetunning Configs}
% We fine-tuned the VideoCrafter-2.0~\cite{chen2023videocrafter1} on our dataset.

\subsection{Video to Music Generation}

To further assess the effectiveness of the video-music pair data, we conduct an extended video-to-music generation task by training VidMuse~\cite{tian2024vidmuse} on an individual music subset of MMTrail-2M. The results, shown in Table~\ref{tab: video-music_gen}, demonstrate the high audio quality and strong audio-visual alignment between the video and the generated music achieved by our dataset.

\begin{table}[h]
  \vspace{-0.5em}
  \caption{Video-to-music generation evaluation results on the MMTrail-Test.}
  \label{tab: video-music_gen}
  \centering
    \setlength\tabcolsep{15pt}%调列距
        \resizebox{0.95\columnwidth}{!}{
  \begin{tabular}{lcccccc}
  \toprule
    Method  & Input & KL$\downarrow$ & ISc$\uparrow$ & FD$\downarrow$ & FAD$\downarrow$ & IB$\uparrow$ \\
    \midrule
    VidMuse~\cite{tian2024vidmuse} & Video & 0.988 & 1.231 & 48.144 & 5.078 & 0.183     \\
    \bottomrule
  \end{tabular}
  }
  \vspace{-0.5em}
\end{table}

% \clearpage

\section{Dataset Details}
\label{sec:sub_details}
Here is one metadata example of our dataset. We introduce the basic information of video and clips in the "basic" tag, including their duration, quality evaluation score, etc.  The useful caption and description are saved in the "scene" tag.
\begin{lstlisting}
[
  {
      'video_id': 'zW1-6V_cN8I',                 # Video ID in MMTrail
      'video_path': 'group_32/zW1-6V_cN8I.mp4',                       # Relative path of the dataset root path
      'video_duration': 1645.52,               # Duration of the video
      'video_resolution': [720, 1280],
      'video_fps': 25.0, 
      'clip_id': 'zW1-6V_cN8I_0000141',           # Clip ID
      'clip_path': 'video_dataset_32/zW1-6V_cN8I_0000141.mp4',          # Relative path of the dataset root path
      'clip_duration': 9.92,            # Duration of the clip itself
      'clip_start_end_idx': [27102, 27350],     # Start frame_id and end frame_id
      'image_quality': 45.510545094807945,      # Image quality score
      'of_score': 6.993135,       # Optical flow score
      'aesthetic_score': [4.515582084655762, 4.1147027015686035, 3.796849250793457], 
      'music_caption_wo_vocal': [{'text': 'This song features a drum machine playing a simple beat. A siren sound is played on the low register. Then, a synth plays a descending lick and the other voice starts rapping. This is followed by a descending run. The mid range of the instruments cannot be heard. This song can be played in a meditation center.', 'time': '0:00-10:00'}],  # Music description of the background music without vocal (human voice).
      'vocal_caption': 'I was just wondering...' # Speech recongitation.
      'frame_caption': ['two people are standing in a room under an umbrella . ', 'a woman in a purple robe standing in front of a man . ', 'a man and a woman dressed in satin robes . '],  # Coca caption of three key frame
      'music_caption': [{'text': 'This music is instrumental. The tempo is medium with a synthesiser arrangement and digital drumming with a lot of vibrato and static. The music is loud, emphatic, youthful, groovy, energetic and pulsating. This music is a Electro Trap.', 'time': '0:00-10:00'}] # Music description of the background music.
      'objects': [' bed', 'Woman', ' wall', ' pink robe', ' pillow'], 
      'background': 'Bedroom', 
      'ocr_score': 0.0, 
      'caption': 'The video shows a woman in a pink robe standing in a room with a bed and a table, captured in a series of keyframes that show her in various poses and expressions.',  # Caption generation from LLaVA and rewrite by LLAMA-13B
      'polish_caption': 'A woman in a pink robe poses and expresses herself in various ways in a room with a bed and a table, capturing her graceful movements and emotive facial expressions.',  # Polished caption generation from LLaVA and rewrite by LLAMA-13B
      'merge_caption': 'In a cozy bedroom setting, a stunning woman adorned in a pink robe gracefully poses and expresses herself, her movements and facial expressions captured in a series of intimate moments. The scene is set against the backdrop of a comfortable bed and a table, with an umbrella standing in a corner of the room. The video features two people standing together under the umbrella, a woman in a purple robe standing confidently in front of a man, and a man and woman dressed in satin robes, all set to an energetic and pulsating electro trap beat with a synthesiser arrangement and digital drumming. The music is loud and emphatic, capturing the youthful and groovy vibe of the video.'# The final description of the video. It is the merge of all above captions, and merged by LLaMA
    }
  }
]
\end{lstlisting}

\clearpage
{\small
\bibliographystyle{plain}
\bibliography{ref}

\begin{thebibliography}{10}

\bibitem{Webvid}
Max Bain, Arsha Nagrani, G{\"u}l Varol, and Andrew Zisserman.
\newblock Frozen in time: A joint video and image encoder for end-to-end retrieval.
\newblock {\em 2021 IEEE/CVF International Conference on Computer Vision (ICCV)}, pages 1708--1718, 2021.

\bibitem{bain2021frozen}
Max Bain, Arsha Nagrani, G{\"u}l Varol, and Andrew Zisserman.
\newblock Frozen in time: A joint video and image encoder for end-to-end retrieval.
\newblock In {\em Proceedings of the IEEE/CVF International Conference on Computer Vision}, pages 1728--1738, 2021.

\bibitem{banerjee-lavie-2005-meteor}
Satanjeev Banerjee and Alon Lavie.
\newblock {METEOR}: An automatic metric for {MT} evaluation with improved correlation with human judgments.
\newblock In Jade Goldstein, Alon Lavie, Chin-Yew Lin, and Clare Voss, editors, {\em Proceedings of the {ACL} Workshop on Intrinsic and Extrinsic Evaluation Measures for Machine Translation and/or Summarization}, pages 65--72, Ann Arbor, Michigan, June 2005. Association for Computational Linguistics.

\bibitem{svd}
Andreas Blattmann, Tim Dockhorn, Sumith Kulal, Daniel Mendelevitch, Maciej Kilian, Dominik Lorenz, Yam Levi, Zion English, Vikram Voleti, Adam Letts, et~al.
\newblock Stable video diffusion: Scaling latent video diffusion models to large datasets.
\newblock {\em arXiv preprint arXiv:2311.15127}, 2023.

\bibitem{blattmann2023align}
Andreas Blattmann, Robin Rombach, Huan Ling, Tim Dockhorn, Seung~Wook Kim, Sanja Fidler, and Karsten Kreis.
\newblock Align your latents: High-resolution video synthesis with latent diffusion models.
\newblock In {\em Proceedings of the IEEE/CVF Conference on Computer Vision and Pattern Recognition}, pages 22563--22575, 2023.

\bibitem{chen2023videocrafter1}
Haoxin Chen, Menghan Xia, Yingqing He, Yong Zhang, Xiaodong Cun, Shaoshu Yang, Jinbo Xing, Yaofang Liu, Qifeng Chen, Xintao Wang, et~al.
\newblock Videocrafter1: Open diffusion models for high-quality video generation.
\newblock {\em arXiv preprint arXiv:2310.19512}, 2023.

\bibitem{chen2020vggsound}
Honglie Chen, Weidi Xie, Andrea Vedaldi, and Andrew Zisserman.
\newblock Vggsound: A large-scale audio-visual dataset.
\newblock In {\em ICASSP 2020-2020 IEEE International Conference on Acoustics, Speech and Signal Processing (ICASSP)}, pages 721--725. IEEE, 2020.

\bibitem{chen2023video}
Jun Chen, Deyao Zhu, Kilichbek Haydarov, Xiang Li, and Mohamed Elhoseiny.
\newblock Video chatcaptioner: Towards the enriched spatiotemporal descriptions.
\newblock {\em arXiv preprint arXiv:2304.04227}, 2023.

\bibitem{panda70m}
Tsai-Shien Chen, Aliaksandr Siarohin, Willi Menapace, Ekaterina Deyneka, Hsiang-wei Chao, Byung~Eun Jeon, Yuwei Fang, Hsin-Ying Lee, Jian Ren, Ming-Hsuan Yang, et~al.
\newblock Panda-70m: Captioning 70m videos with multiple cross-modality teachers.
\newblock {\em arXiv preprint arXiv:2402.19479}, 2024.

\bibitem{chi2024m2chat}
Xiaowei Chi, Yijiang Liu, Zhengkai Jiang, Rongyu Zhang, Ziyi Lin, Renrui Zhang, Peng Gao, Chaoyou Fu, Shanghang Zhang, Qifeng Liu, et~al.
\newblock Chatillusion: Efficient-aligning interleaved generation ability with visual instruction model.
\newblock {\em arXiv preprint arXiv:2311.17963}, 2023.

\bibitem{copet2024simple}
Jade Copet, Felix Kreuk, Itai Gat, Tal Remez, David Kant, Gabriel Synnaeve, Yossi Adi, and Alexandre D{\'e}fossez.
\newblock Simple and controllable music generation.
\newblock {\em Advances in Neural Information Processing Systems}, 36, 2024.

\bibitem{musiccap}
SeungHeon Doh, Keunwoo Choi, Jongpil Lee, and Juhan Nam.
\newblock Lp-musiccaps: Llm-based pseudo music captioning.
\newblock {\em arXiv preprint arXiv:2307.16372}, 2023.

\bibitem{doh2023lp}
SeungHeon Doh, Keunwoo Choi, Jongpil Lee, and Juhan Nam.
\newblock Lp-musiccaps: Llm-based pseudo music captioning.
\newblock {\em arXiv preprint arXiv:2307.16372}, 2023.

\bibitem{dosovitskiy2021image}
Alexey Dosovitskiy, Lucas Beyer, Alexander Kolesnikov, Dirk Weissenborn, Xiaohua Zhai, Thomas Unterthiner, Mostafa Dehghani, Matthias Minderer, Georg Heigold, Sylvain Gelly, Jakob Uszkoreit, and Neil Houlsby.
\newblock An image is worth 16x16 words: Transformers for image recognition at scale, 2021.

\bibitem{gemmeke2017audio}
Jort~F Gemmeke, Daniel~PW Ellis, Dylan Freedman, Aren Jansen, Wade Lawrence, R~Channing Moore, Manoj Plakal, and Marvin Ritter.
\newblock Audio set: An ontology and human-labeled dataset for audio events.
\newblock In {\em 2017 IEEE international conference on acoustics, speech and signal processing (ICASSP)}, pages 776--780. IEEE, 2017.

\bibitem{he2024llmsmeetmultimodalgeneration}
Yingqing He, Zhaoyang Liu, Jingye Chen, Zeyue Tian, Hongyu Liu, Xiaowei Chi, Runtao Liu, Ruibin Yuan, Yazhou Xing, Wenhai Wang, et~al.
\newblock Llms meet multimodal generation and editing: A survey.
\newblock {\em arXiv preprint arXiv:2405.19334}, 2024.

\bibitem{animate-a-story}
Yingqing He, Menghan Xia, Haoxin Chen, Xiaodong Cun, Yuan Gong, Jinbo Xing, Yong Zhang, Xintao Wang, Chao Weng, Ying Shan, et~al.
\newblock Animate-a-story: Storytelling with retrieval-augmented video generation.
\newblock {\em arXiv preprint arXiv:2307.06940}, 2023.

\bibitem{he2022lvdm}
Yingqing He, Tianyu Yang, Yong Zhang, Ying Shan, and Qifeng Chen.
\newblock Latent video diffusion models for high-fidelity long video generation.
\newblock {\em arXiv preprint arXiv:2211.13221}, 2022.

\bibitem{henschel2024streamingt2v}
Roberto Henschel, Levon Khachatryan, Daniil Hayrapetyan, Hayk Poghosyan, Vahram Tadevosyan, Zhangyang Wang, Shant Navasardyan, and Humphrey Shi.
\newblock Streamingt2v: Consistent, dynamic, and extendable long video generation from text.
\newblock {\em arXiv preprint arXiv:2403.14773}, 2024.

\bibitem{ho2022imagen-video}
Jonathan Ho, William Chan, Chitwan Saharia, Jay Whang, Ruiqi Gao, Alexey Gritsenko, Diederik~P Kingma, Ben Poole, Mohammad Norouzi, David~J Fleet, et~al.
\newblock Imagen video: High definition video generation with diffusion models.
\newblock {\em arXiv preprint arXiv:2210.02303}, 2022.

\bibitem{vdm}
Jonathan Ho, Tim Salimans, Alexey Gritsenko, William Chan, Mohammad Norouzi, and David~J Fleet.
\newblock Video diffusion models.
\newblock {\em Advances in Neural Information Processing Systems}, 35:8633--8646, 2022.

\bibitem{huang2023vbench}
Ziqi Huang, Yinan He, Jiashuo Yu, Fan Zhang, Chenyang Si, Yuming Jiang, Yuanhan Zhang, Tianxing Wu, Qingyang Jin, Nattapol Chanpaisit, Yaohui Wang, Xinyuan Chen, Limin Wang, Dahua Lin, Yu~Qiao, and Ziwei Liu.
\newblock {VBench}: Comprehensive benchmark suite for video generative models.
\newblock In {\em Proceedings of the IEEE/CVF Conference on Computer Vision and Pattern Recognition}, 2024.

\bibitem{jin2023chatunivi}
Peng Jin, Ryuichi Takanobu, Caiwan Zhang, Xiaochun Cao, and Li~Yuan.
\newblock Chat-univi: Unified visual representation empowers large language models with image and video understanding.
\newblock {\em arXiv preprint arXiv:2311.08046}, 2023.

\bibitem{kilgour2018fr}
Kevin Kilgour, Mauricio Zuluaga, Dominik Roblek, and Matthew Sharifi.
\newblock Fr$\backslash$'echet audio distance: A metric for evaluating music enhancement algorithms.
\newblock {\em arXiv preprint arXiv:1812.08466}, 2018.

\bibitem{kondratyuk2023videopoet}
Dan Kondratyuk, Lijun Yu, Xiuye Gu, Jos{\'e} Lezama, Jonathan Huang, Rachel Hornung, Hartwig Adam, Hassan Akbari, Yair Alon, Vighnesh Birodkar, et~al.
\newblock Videopoet: A large language model for zero-shot video generation.
\newblock {\em arXiv preprint arXiv:2312.14125}, 2023.

\bibitem{kong2020panns}
Qiuqiang Kong, Yin Cao, Turab Iqbal, Yuxuan Wang, Wenwu Wang, and Mark~D Plumbley.
\newblock Panns: Large-scale pretrained audio neural networks for audio pattern recognition.
\newblock {\em IEEE/ACM Transactions on Audio, Speech, and Language Processing}, 28:2880--2894, 2020.

\bibitem{ActivityNetCaptions}
Ranjay Krishna, Kenji Hata, Frederic Ren, Li~Fei-Fei, and Juan~Carlos Niebles.
\newblock Dense-captioning events in videos.
\newblock {\em 2017 IEEE International Conference on Computer Vision (ICCV)}, pages 706--715, 2017.

\bibitem{li2024mgm}
Yanwei Li, Yuechen Zhang, Chengyao Wang, Zhisheng Zhong, Yixin Chen, Ruihang Chu, Shaoteng Liu, and Jiaya Jia.
\newblock Mini-gemini: Mining the potential of multi-modality vision language models.
\newblock {\em arXiv:2403.18814}, 2023.

\bibitem{videollava}
Bin Lin, Bin Zhu, Yang Ye, Munan Ning, Peng Jin, and Li~Yuan.
\newblock Video-llava: Learning united visual representation by alignment before projection.
\newblock {\em arXiv preprint arXiv:2311.10122}, 2023.

\bibitem{lin-och-2004-automatic}
Chin-Yew Lin and Franz~Josef Och.
\newblock Automatic evaluation of machine translation quality using longest common subsequence and skip-bigram statistics.
\newblock In {\em Proceedings of the 42nd Annual Meeting of the Association for Computational Linguistics ({ACL}-04)}, pages 605--612, Barcelona, Spain, July 2004.

\bibitem{lin2023videodirectorgpt}
Han Lin, Abhay Zala, Jaemin Cho, and Mohit Bansal.
\newblock Videodirectorgpt: Consistent multi-scene video generation via llm-guided planning.
\newblock {\em arXiv preprint arXiv:2309.15091}, 2023.

\bibitem{lin2019tsm}
Ji~Lin, Chuang Gan, and Song Han.
\newblock Tsm: Temporal shift module for efficient video understanding.
\newblock In {\em Proceedings of the IEEE/CVF international conference on computer vision}, pages 7083--7093, 2019.

\bibitem{lin2022swinbert}
Kevin Lin, Linjie Li, Chung-Ching Lin, Faisal Ahmed, Zhe Gan, Zicheng Liu, Yumao Lu, and Lijuan Wang.
\newblock Swinbert: End-to-end transformers with sparse attention for video captioning.
\newblock In {\em Proceedings of the IEEE/CVF Conference on Computer Vision and Pattern Recognition}, pages 17949--17958, 2022.

\bibitem{lin2015microsoft}
Tsung-Yi Lin, Michael Maire, Serge Belongie, Lubomir Bourdev, Ross Girshick, James Hays, Pietro Perona, Deva Ramanan, C.~Lawrence Zitnick, and Piotr Dollár.
\newblock Microsoft coco: Common objects in context, 2015.

\bibitem{liu2024llavanext}
Haotian Liu, Chunyuan Li, Yuheng Li, Bo~Li, Yuanhan Zhang, Sheng Shen, and Yong~Jae Lee.
\newblock Llava-next: Improved reasoning, ocr, and world knowledge, January 2024.

\bibitem{liu2021tam}
Zhaoyang Liu, Limin Wang, Wayne Wu, Chen Qian, and Tong Lu.
\newblock Tam: Temporal adaptive module for video recognition.
\newblock In {\em Proceedings of the IEEE/CVF international conference on computer vision}, pages 13708--13718, 2021.

\bibitem{long2024videodrafter}
Fuchen Long, Zhaofan Qiu, Ting Yao, and Tao Mei.
\newblock Videodrafter: Content-consistent multi-scene video generation with llm.
\newblock {\em arXiv preprint arXiv:2401.01256}, 2024.

\bibitem{maaz2023videochatgpt}
Muhammad Maaz, Hanoona Rasheed, Salman Khan, and Fahad~Shahbaz Khan.
\newblock Video-chatgpt: Towards detailed video understanding via large vision and language models, 2023.

\bibitem{HowTo100M}
Antoine Miech, Dimitri Zhukov, Jean-Baptiste Alayrac, Makarand Tapaswi, Ivan Laptev, and Josef Sivic.
\newblock Howto100m: Learning a text-video embedding by watching hundred million narrated video clips.
\newblock {\em 2019 IEEE/CVF International Conference on Computer Vision (ICCV)}, pages 2630--2640, 2019.

\bibitem{VideoCC3M}
Arsha Nagrani, Paul~Hongsuck Seo, Bryan Seybold, Anja Hauth, Santiago Man{\'e}n, Chen Sun, and Cordelia Schmid.
\newblock Learning audio-video modalities from image captions.
\newblock In {\em European Conference on Computer Vision}, 2022.

\bibitem{papineni-etal-2002-bleu}
Kishore Papineni, Salim Roukos, Todd Ward, and Wei-Jing Zhu.
\newblock {B}leu: a method for automatic evaluation of machine translation.
\newblock In Pierre Isabelle, Eugene Charniak, and Dekang Lin, editors, {\em Proceedings of the 40th Annual Meeting of the Association for Computational Linguistics}, pages 311--318, Philadelphia, Pennsylvania, USA, July 2002. Association for Computational Linguistics.

\bibitem{LSMDC}
Anna Rohrbach, Atousa Torabi, Marcus Rohrbach, Niket Tandon, Christopher~Joseph Pal, H.~Larochelle, Aaron~C. Courville, and Bernt Schiele.
\newblock Movie description.
\newblock {\em International Journal of Computer Vision}, 123:94 -- 120, 2016.

\bibitem{demucs}
Simon Rouard, Francisco Massa, and Alexandre D{\'e}fossez.
\newblock Hybrid transformers for music source separation.
\newblock In {\em ICASSP 23}, 2023.

\bibitem{How2}
Ramon Sanabria, Ozan Caglayan, Shruti Palaskar, Desmond Elliott, Lo{\"i}c Barrault, Lucia Specia, and Florian Metze.
\newblock How2: A large-scale dataset for multimodal language understanding.
\newblock {\em ArXiv}, abs/1811.00347, 2018.

\bibitem{schuhmann2021laion400m}
Christoph Schuhmann, Richard Vencu, Romain Beaumont, Robert Kaczmarczyk, Clayton Mullis, Aarush Katta, Theo Coombes, Jenia Jitsev, and Aran Komatsuzaki.
\newblock Laion-400m: Open dataset of clip-filtered 400 million image-text pairs, 2021.

\bibitem{singer2022make-a-video}
Uriel Singer, Adam Polyak, Thomas Hayes, Xi~Yin, Jie An, Songyang Zhang, Qiyuan Hu, Harry Yang, Oron Ashual, Oran Gafni, et~al.
\newblock Make-a-video: Text-to-video generation without text-video data.
\newblock {\em arXiv preprint arXiv:2209.14792}, 2022.

\bibitem{song2023moviechat}
Enxin Song, Wenhao Chai, Guanhong Wang, Yucheng Zhang, Haoyang Zhou, Feiyang Wu, Xun Guo, Tian Ye, Yan Lu, Jenq-Neng Hwang, et~al.
\newblock Moviechat: From dense token to sparse memory for long video understanding.
\newblock {\em arXiv preprint arXiv:2307.16449}, 2023.

\bibitem{soomro2012ucf101}
Khurram Soomro, Amir~Roshan Zamir, and Mubarak Shah.
\newblock Ucf101: A dataset of 101 human actions classes from videos in the wild.
\newblock {\em arXiv preprint arXiv:1212.0402}, 2012.

\bibitem{WTS70M}
Jonathan~C. Stroud, David~A. Ross, Chen Sun, Jia Deng, Rahul Sukthankar, and Cordelia Schmid.
\newblock Learning video representations from textual web supervision.
\newblock {\em ArXiv}, abs/2007.14937, 2020.

\bibitem{tian2024vidmuse}
Zeyue Tian, Zhaoyang Liu, Ruibin Yuan, Jiahao Pan, Xiaoqiang Huang, Qifeng Liu, Xu~Tan, Qifeng Chen, Wei Xue, and Yike Guo.
\newblock Vidmuse: A simple video-to-music generation framework with long-short-term modeling.
\newblock {\em arXiv preprint arXiv:2406.04321}, 2024.

\bibitem{touvron2023llama2}
Hugo Touvron, Louis Martin, Kevin Stone, Peter Albert, Amjad Almahairi, Yasmine Babaei, Nikolay Bashlykov, Soumya Batra, Prajjwal Bhargava, Shruti Bhosale, et~al.
\newblock Llama 2: Open foundation and fine-tuned chat models.
\newblock {\em arXiv preprint arXiv:2307.09288}, 2023.

\bibitem{vedantam2015cider}
Ramakrishna Vedantam, C.~Lawrence Zitnick, and Devi Parikh.
\newblock Cider: Consensus-based image description evaluation, 2015.

\bibitem{wang2023gen}
Fu-Yun Wang, Wenshuo Chen, Guanglu Song, Han-Jia Ye, Yu~Liu, and Hongsheng Li.
\newblock Gen-l-video: Multi-text to long video generation via temporal co-denoising.
\newblock {\em arXiv preprint arXiv:2305.18264}, 2023.

\bibitem{HD-VG-130M}
Wenjing Wang, Huan Yang, Zixi Tuo, Huiguo He, Junchen Zhu, Jianlong Fu, and Jiaying Liu.
\newblock Videofactory: Swap attention in spatiotemporal diffusions for text-to-video generation.
\newblock {\em ArXiv}, abs/2305.10874, 2023.

\bibitem{VaTeX}
Xin~Eric Wang, Jiawei Wu, Junkun Chen, Lei Li, Yuan fang Wang, and William~Yang Wang.
\newblock Vatex: A large-scale, high-quality multilingual dataset for video-and-language research.
\newblock {\em 2019 IEEE/CVF International Conference on Computer Vision (ICCV)}, pages 4580--4590, 2019.

\bibitem{InternVid}
Yi~Wang, Yinan He, Yizhuo Li, Kunchang Li, Jiashuo Yu, Xin~Jian Ma, Xinyuan Chen, Yaohui Wang, Ping Luo, Ziwei Liu, Yali Wang, Limin Wang, and Y.~Qiao.
\newblock Internvid: A large-scale video-text dataset for multimodal understanding and generation.
\newblock {\em ArXiv}, abs/2307.06942, 2023.

\bibitem{wu2021star}
Bo~Wu, Shoubin Yu, Zhenfang Chen, Joshua~B Tenenbaum, and Chuang Gan.
\newblock Star: A benchmark for situated reasoning in real-world videos.
\newblock In {\em Thirty-fifth conference on neural information processing systems datasets and benchmarks track (Round 2)}, 2021.

\bibitem{wu2019long}
Chao-Yuan Wu, Christoph Feichtenhofer, Haoqi Fan, Kaiming He, Philipp Krahenbuhl, and Ross Girshick.
\newblock Long-term feature banks for detailed video understanding.
\newblock In {\em Proceedings of the IEEE/CVF Conference on Computer Vision and Pattern Recognition}, pages 284--293, 2019.

\bibitem{xiang2024pandora}
Jiannan Xiang, Guangyi Liu, Yi~Gu, Qiyue Gao, Yuting Ning, Yuheng Zha, Zeyu Feng, Tianhua Tao, Shibo Hao, Yemin Shi, Zhengzhong Liu, Eric~P. Xing, and Zhiting Hu.
\newblock Pandora: Towards general world model with natural language actions and video states.
\newblock {\em arXiv preprint arXiv:2406.09455}, 2024.

\bibitem{MSR-VTT}
Jun Xu, Tao Mei, Ting Yao, and Yong Rui.
\newblock Msr-vtt: A large video description dataset for bridging video and language.
\newblock {\em 2016 IEEE Conference on Computer Vision and Pattern Recognition (CVPR)}, pages 5288--5296, 2016.

\bibitem{HD-VILA}
Hongwei Xue, Tiankai Hang, Yanhong Zeng, Yuchong Sun, Bei Liu, Huan Yang, Jianlong Fu, and Baining Guo.
\newblock Advancing high-resolution video-language representation with large-scale video transcriptions.
\newblock {\em 2022 IEEE/CVF Conference on Computer Vision and Pattern Recognition (CVPR)}, pages 5026--5035, 2021.

\bibitem{yang2022learning}
Antoine Yang, Antoine Miech, Josef Sivic, Ivan Laptev, and Cordelia Schmid.
\newblock Learning to answer visual questions from web videos.
\newblock {\em arXiv preprint arXiv:2205.05019}, 2022.

\bibitem{yang2022zero}
Antoine Yang, Antoine Miech, Josef Sivic, Ivan Laptev, and Cordelia Schmid.
\newblock Zero-shot video question answering via frozen bidirectional language models.
\newblock {\em Advances in Neural Information Processing Systems}, 35:124--141, 2022.

\bibitem{Yu2022CoCa}
Jiahui Yu, Zirui Wang, Vijay Vasudevan, Legg Yeung, Mojtaba Seyedhosseini, and Yonghui Wu.
\newblock Coca: Contrastive captioners are image-text foundation models.
\newblock {\em Trans. Mach. Learn. Res.}, 2022, 2022.

\bibitem{yuan2024chatmusician}
Ruibin Yuan, Hanfeng Lin, Yi~Wang, Zeyue Tian, Shangda Wu, Tianhao Shen, Ge~Zhang, Yuhang Wu, Cong Liu, Ziya Zhou, et~al.
\newblock Chatmusician: Understanding and generating music intrinsically with llm.
\newblock {\em arXiv preprint arXiv:2402.16153}, 2024.

\bibitem{YT-Temporal-180M}
Rowan Zellers, Ximing Lu, Jack Hessel, Youngjae Yu, Jae~Sung Park, Jize Cao, Ali Farhadi, and Yejin Choi.
\newblock Merlot: Multimodal neural script knowledge models.
\newblock In {\em Neural Information Processing Systems}, 2021.

\bibitem{videollama}
Hang Zhang, Xin Li, and Lidong Bing.
\newblock Video-llama: An instruction-tuned audio-visual language model for video understanding.
\newblock {\em arXiv preprint arXiv:2306.02858}, 2023.

\bibitem{zhang2023vision}
Qiming Zhang, Jing Zhang, Yufei Xu, and Dacheng Tao.
\newblock Vision transformer with quadrangle attention.
\newblock {\em arXiv preprint arXiv:2303.15105}, 2023.

\bibitem{zhang2020bertscore}
Tianyi Zhang, Varsha Kishore, Felix Wu, Kilian~Q. Weinberger, and Yoav Artzi.
\newblock Bertscore: Evaluating text generation with bert, 2020.

\bibitem{zhao2022towards}
Minyi Zhao, Bingjia Li, Jie Wang, Wanqing Li, Wenjing Zhou, Lan Zhang, Shijie Xuyang, Zhihang Yu, Xinkun Yu, Guangze Li, et~al.
\newblock Towards video text visual question answering: benchmark and baseline.
\newblock {\em Advances in Neural Information Processing Systems}, 35:35549--35562, 2022.

\bibitem{zheng2023judging}
Lianmin Zheng, Wei-Lin Chiang, Ying Sheng, Siyuan Zhuang, Zhanghao Wu, Yonghao Zhuang, Zi~Lin, Zhuohan Li, Dacheng Li, Eric.~P Xing, Hao Zhang, Joseph~E. Gonzalez, and Ion Stoica.
\newblock Judging llm-as-a-judge with mt-bench and chatbot arena, 2023.

\bibitem{zhou2022magicvideo}
Daquan Zhou, Weimin Wang, Hanshu Yan, Weiwei Lv, Yizhe Zhu, and Jiashi Feng.
\newblock Magicvideo: Efficient video generation with latent diffusion models.
\newblock {\em arXiv preprint arXiv:2211.11018}, 2022.

\bibitem{zhu2023minigpt4}
Deyao Zhu, Jun Chen, Xiaoqian Shen, Xiang Li, and Mohamed Elhoseiny.
\newblock Minigpt-4: Enhancing vision-language understanding with advanced large language models.
\newblock {\em arXiv preprint arXiv:2304.10592}, 2023.

\bibitem{zhu2023minigpt}
Deyao Zhu, Jun Chen, Xiaoqian Shen, Xiang Li, and Mohamed Elhoseiny.
\newblock Minigpt-4: Enhancing vision-language understanding with advanced large language models.
\newblock {\em arXiv preprint arXiv:2304.10592}, 2023.

\end{thebibliography}
}
%%%%%%%%%%%%%%%%%%%%%%%%%%%%%%%%%%%%%%%%%%%%%%%%%%%%%%%%%%%%
\clearpage
% \input{sec/6_checklist}
% \appendix

\end{document}